\documentclass[referee, sn-mathphys-num]{sn-jnl}

\usepackage{graphicx}%
\usepackage{multirow}%
\usepackage{amsmath,amssymb,amsfonts}%
\usepackage{amsthm}%
\usepackage{siunitx}%
\usepackage{mathrsfs}%
\usepackage[title]{appendix}%
\usepackage{xcolor}%
\usepackage{textcomp}%
\usepackage{manyfoot}%
\usepackage{booktabs}%
\usepackage{algorithm}%
\usepackage{algorithmicx}%
\usepackage{algpseudocode}%
\usepackage{listings}%
\usepackage{bm}%
\usepackage{url}%
\usepackage[normalem]{ulem}

\newcommand{\figref}[1]{{Fig.~\ref{#1}}}
\newcommand{\tabref}[1]{{Tab.~\ref{#1}}}
\newcommand{\equref}[1]{{equation~(\ref{#1})}}

\makeatletter
\newcommand{\useextendeddatafiglabel}{%
  \renewcommand{\fnum@figure}{Extended Data Fig.~\thefigure}%
}

\newcommand{\usenormalfiglabel}{%
  \renewcommand{\fnum@figure}{Fig.~\thefigure}%
}

\newcommand{\useextendeddatatablabel}{%
  \renewcommand{\fnum@table}{Extended Data Table ~\thetable}%
}
\newcommand{\usenormaltablabel}{%
  \renewcommand{\fnum@table}{Fig.~\thetable}%
}

\renewcommand\thesection{}
\renewcommand\thesubsection{\arabic{subsection}}

\makeatother

\begin{document} 


\title[Article Title]{Self-assembling Modular Aerial Robot for\\Versatile Aerial Tasks}



\author[1]{\fnm{Junichiro} \sur{Sugihara}}

\author[1]{\fnm{Masaki} \sur{Kitagawa}}

\author[1]{\fnm{Jinjie} \sur{Li}}
\author[1]{\fnm{Yunong} \sur{Li}}

\author[2]{\fnm{Takuzumi} \sur{Nishio}}
\author[2]{\fnm{Kei} \sur{Okada}}

\author*[1]{\fnm{Moju} \sur{Zhao}}\email{chou@dragon.t.u-tokyo.ac.jp}

\affil*[1]{\orgdiv{Department of Mechanical Engineering}, \orgname{The University of Tokyo},\state{Tokyo}, \country{Japan}}
\affil[2]{\orgdiv{Department of Mechano Informatics}, \orgname{The University of Tokyo},\state{Tokyo}, \country{Japan}}

\abstract{
Multirotor aerial robots excel at maneuvering in three-dimensional space, and recent advances enable nimble navigation in cluttered and confined environments, especially for small airframes. 
\
By contrast, platforms built for high-altitude work tend to be larger to deliver high thrust for stable physical interaction with the environment. However, these conflicting design requirements create a long-standing trade-off between nimble navigation and robust aerial manipulation. 
\
Here, we present LEGION units, which are reconfigurable modular aerial robots capable of in-flight self-assembly for cooperative manipulation, drawing inspiration from the self-organized collectives formed by ants. 
\
Each unit retains nimble maneuverability while joint-equipped docking interfaces at both ends enable end-to-end self-assembly into a flying manipulator.
\
We show that multiple units autonomously dock in flight; once latched, they maintain a zero-clearance interlock by controlling the contact force and torque, enabling reliable aggregation and articulated motion even outdoors.
\
We further show that self-reconfigurability enables morphological switching between nimble individual flight and collective articulated manipulation, while realizing core in-flight manipulation primitives including pushing, pulling, rotating, grasping, and carrying. 
\
LEGION’s self-organization enables aerial robots, especially in swarms, to shift from passive observers to active participants in their environment, broadening the scope of aerial physical interaction.
}

\maketitle

\noindent
\section*{Introduction}

\begin{figure}
\centering
\includegraphics[width=0.9\textwidth]{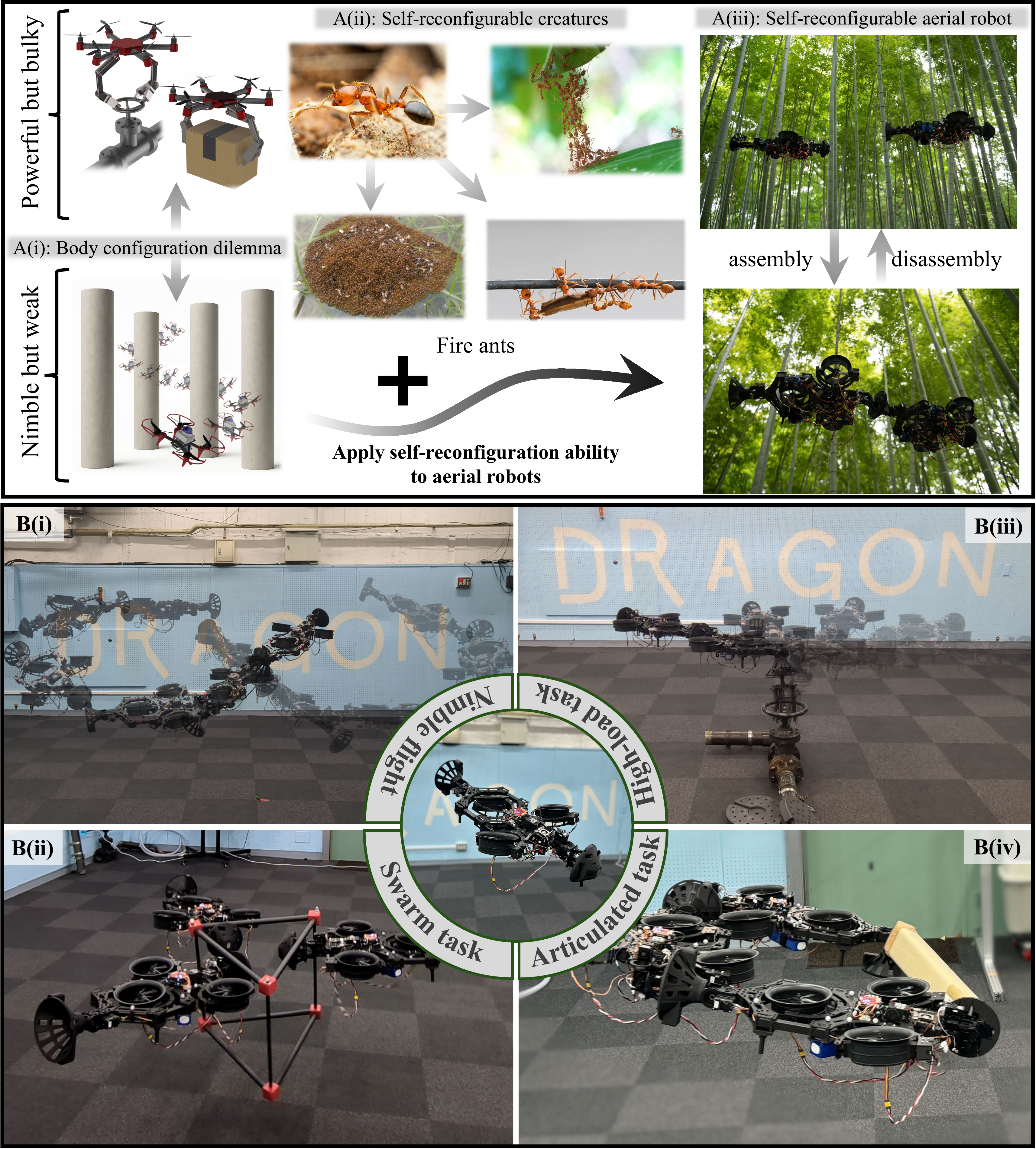}
 \caption{\textbf{Overview of the background and proposed aerial modular robotic platform, LEGION.} 
 (\textbf{A}) Conceptual background of this study. (i) The body configuration dilemma in aerial robots. (ii) Self-reconfiguration capability of fire ants. (iii) Proposed solution: a self-reconfigurable aerial robot.
(\textbf{B}) Multiple tasks achieved by LEGIONs in different configurations. (i) Nimble flight in the separated state.  (ii) Cooperative manipulation tasks performed by the swarm. (iii) Manipulation task requiring high end-effector torque. (iv) Manipulation task requiring articulated joint motion. (credit: fire ants \cite{fire-ant, ant-raft, ant-bridge, ant-carrying})}
 \label{fig:overview}
\end{figure}

Multirotor aerial robots offer exceptional mobility, enabling diverse applications such as cinematography, inspection, and entertainment  \cite{cinematography-2020-JFR, subterran-2022-SR, drone-show-2025-ICRA}.
\
Their compact version shows incredible nimbleness in cluttered environments based on autonomous obstacle avoidance \cite{min-snap-2011-ICRA, freestyle-2025-SR, safe-nav-2025-SR}, and swarm systems have also been realized \cite{dcad-2020-RAL, omni-swarm-2022-TRO, swarm-2022-SR}.
The agility of aerial robots then yields a new sport called drone racing \cite{drone-race-2024-TRO}, and recent progress achieves autonomy comparable to human pilots \cite{swift-2023-nature}. 
Another notable potential of multirotor aerial robot is aerial manipulation, where a robotic hand or arm is deployed into the body \cite{am-survey-2022-TRO}. This coupled system can perform complex three-dimensional manipulation tasks at hard-to-reach locations, including aerial grasping \cite{grasp-2011-IROS, soft-gripper-2024-SA, fast-grasp-2024-npjr} and dexterous manipulation \cite{am-prototype-2014-RAM, aeroarms-2018-RAM, aam-2022-Nature, millimeter-am-2024-TRO, flyingtoolbox-2025-Nature}. These aerial physical interaction tasks require the multirotor body to be larger to deliver high thrust.
Thus, smaller aerial robots offer high mobility but lack the power required for heavy manipulation tasks, while manipulation-oriented bulky aerial robots sacrifice mobility \cite{aerila-robot-2012-IJRR}. These conflicting body configurations create a long-standing trade-off between nimble navigation and aerial manipulation (\figref{fig:overview}A(i)).
\

One possible solution to this trade-off is the morphing mechanism that enables expanding and contracting the body, which provides advanced navigation in confined environments; however, the manipulation ability is limited to simple grasping tasks \cite{deform-drone-2017-IROS, foldable-drone-2019-RAL}. Then, articulated aerial robots that distribute rotors across multiple links are proposed to enable the snake-like maneuvering to squeeze through small gaps or openings in midair \cite{dragon-squeeze-2020-RAL, chain-quadorotor-2020-ICRA}, while also providing the manipulator-like capabilities \cite{lasdra-2018-ICRA, hydrus-2017-IJRR, dargon-valve-2022-RAL, dragon-2023-IJRR, perching-arm-2024-TRO}. However, the snake-like motion by these robots cannot completely solve the collision problem in more confined environments, since the links are tightly connected. 
This limitation leads to the concept of separable aerial robots  \cite{dfa-2012-IJRR, modquad-2018-ICRA,modular-quadrotor-2024-RAL, trady-2023-AIS,beatle-2024-TMech,dockable-multirotor-2025-Access}. Although individual unit has the nimble---even agile---mobility, they can only form a larger, bulky rigid body that lacks the aerial manipulation capability. Therefore, a true modularity that can perform reconfiguration after self-assembly is expected to gain both the individual nimbleness and the collective manipulability.


Here, drawing inspiration from ants that can aggregate into a flexible cargo to transport their prey \cite{transport-2018-NatureP}  and into a flexible bridge to traverse gaps \cite{army-ants-2015-NAS} (\figref{fig:overview}A(ii)), we propose a modular aerial robot called LEGION (Linkable aErial robot with GImballed rotors and jOiNts) capable of self-organization to perform cooperative manipulation  (\figref{fig:overview}A(iii)). Our modular aerial robots can perform nimble flight individually (\figref{fig:overview}B(i)) and then assemble into a larger reconfigurable structure for versatile aerial manipulation tasks(\figref{fig:overview}B(ii-iv)).

To achieve this unique modularity, each robot is equipped with a pair of docking interfaces---each incorporating a joint---at its two ends, allowing the system to dock and then transition into an aerial manipulator capable of articulated motion, thereby enabling task primitives such as pushing, rotating, grasping, and transporting.
\
Compared with modular robots operating on the ground  \cite{modular-robot-survey-2025, snail-robot-2024-NC}, inflight precise docking is more challenging because aerial robots are floating systems and are easily influenced by mutual disturbances.
\
Even if inflight docking were achievable, previous work  \cite{dfa-2012-IJRR,h-mod-quad-2021-ICRA,mars-2025-ICRA} still ignores the interaction force and torque---i.e., the contact wrench---between interlocked modules during flight, as the aggregated structure is treated as a single rigid body. For an articulated configuration, however, the inter-module contact wrench must be accurately estimated and controlled.



To address these two challenges, we designed three core methods: (1) detachable magnetic--mechanical hybrid docking mechanism; (2) robust inflight docking planning technique; and (3) contact wrench estimation and control method.
These methods enable the LEGION swarm to self-assemble and morph during flight in a highly decentralized manner. Our docking system also enables fully autonomous outdoor self-assembly and morphing using only onboard sensors for localization and external wrench estimation.
\
Furthermore, it is demonstrated that multiple LEGIONs can cooperatively perform various manipulation tasks including pushing, pulling, rotating, grasping, and carrying, which can be considered as five manipulation primitives in the aerial domain.
By resolving the maneuverability--manipulability paradox, LEGION's self-organization capability opens the way to new deployment opportunities for aerial robots across a wide spectrum of applications involving physical interaction.

\section*{Results}
\subsection*{Self-assembling aerial robot}
\begin{figure}
 \centering
 \includegraphics[width=1.0\textwidth]{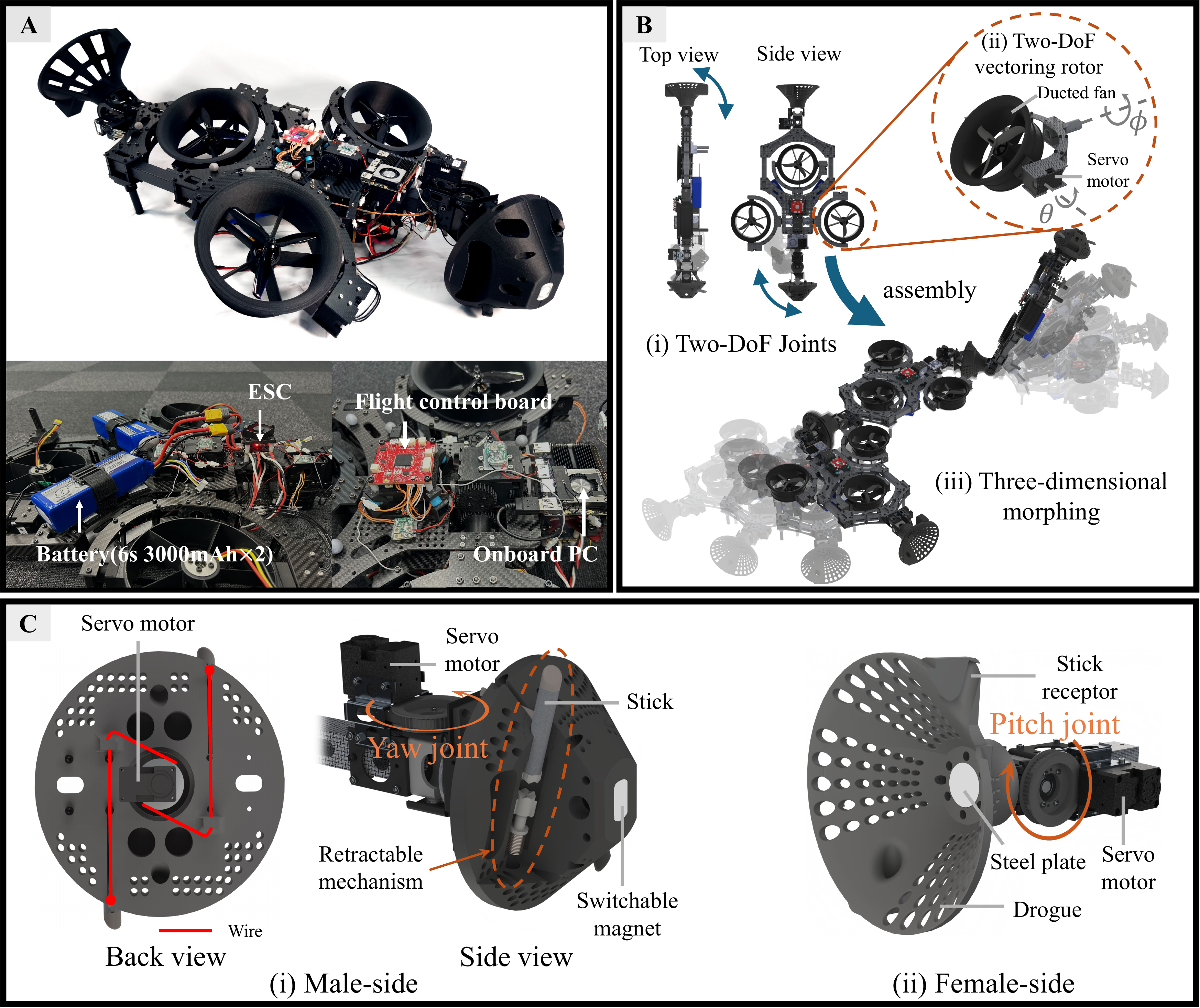}
 \caption{\textbf{Overall mechanical design of LEGION.} 
 (\textbf{A}) Real prototype of the proposed LEGION module. 
 (\textbf{B}) Mechanical configuration of the airframe. (i) Joint configuration. (ii) Rotor configuration. (iii) Example of joint motion in a three-module assembled configuration. 
 (\textbf{C}) Detailed design of the docking mechanism, showing (i) the male side and (ii) the female side.}
\label{fig:hardware}
\end{figure}

\begin{figure}[!b]
 \centering
 \includegraphics[width=0.85\textwidth]{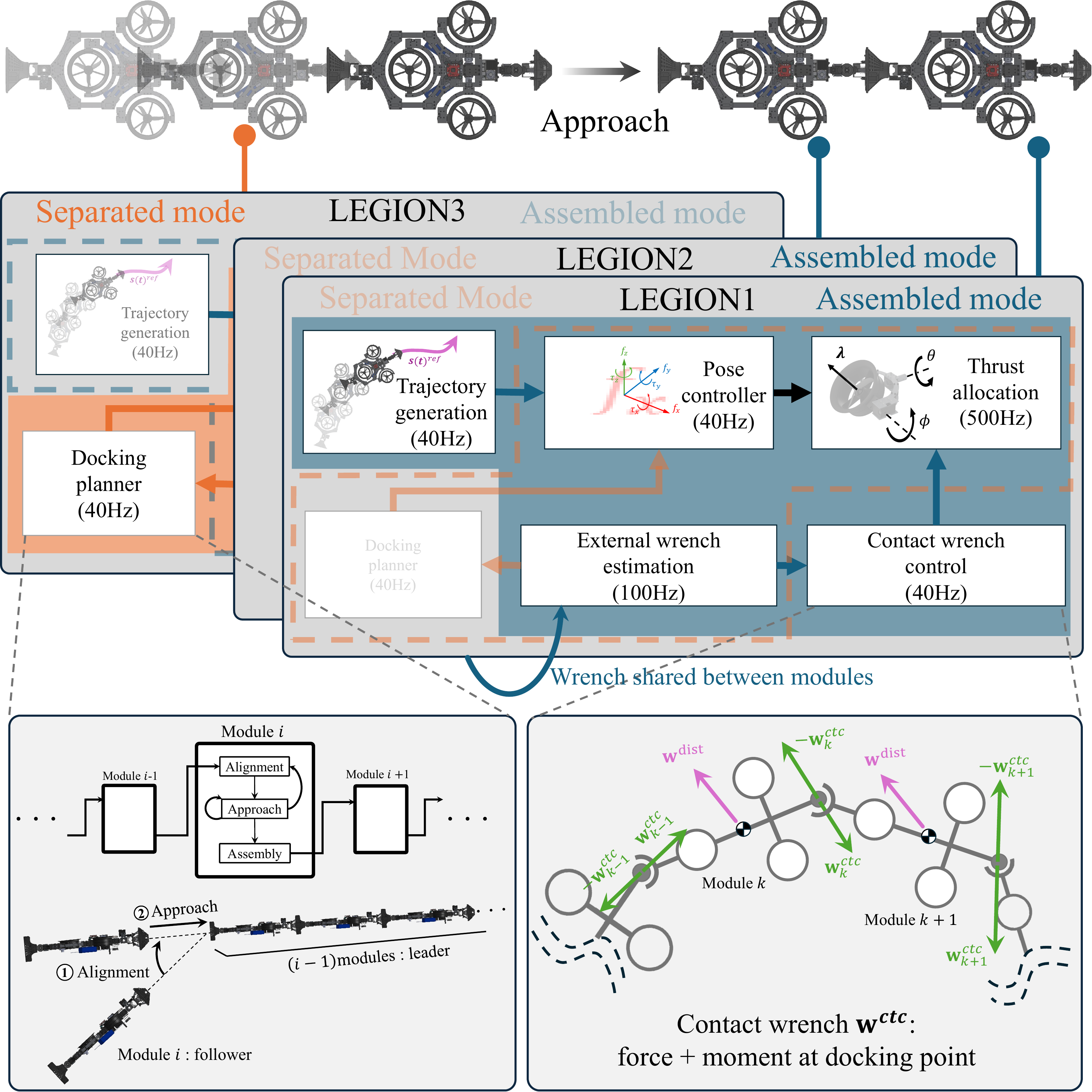}
 \caption{\textbf{Decentralized control architecture of LEGION}. Each LEGION module independently performs the flight control. The control framework operating inside each module consists of multiple control components, and the combination switches depending on whether the module is in the separated (blue area) or assembled  (orange area) mode. Inflight docking between modules is fully autonomous and governed by a state machine composed of three states: alignment, approach and assembly. During flight in the assembled mode, stable collective flight is achieved by having each module cooperatively control the contact wrench---comprising force and torque---at the docking point. This wrench is estimated using a sensor-less observer embedded in each module, and is shared with other modules to enable coordinated control. Details of each component are provided in  ``Materials and Methods''.}
\label{fig:control_framework}
\end{figure}

An overview of the mechanical configuration of the LEGION module is shown in \figref{fig:hardware}.
The two key components of the module are vectoring rotors (\figref{fig:hardware}B(ii)) and docking mechanism (\figref{fig:hardware}C).
\
\
Compared with a four-rotor configuration \cite{gripper-modquad-2018-ICRA}, the configuration by three-rotor can offer a larger morphing range while aggregating.
Each rotor can vector its thrust about two perpendicular rotation axes and thus counteract gravity at arbitrary poses.
\
Based on the fully onboard flight control system (\figref{fig:control_framework} and ``Flight control'' in Materials and Methods), trajectory tracking test was performed, demonstrating omni-directional maneuverability.
The average tracking pose errors of 0.15 m and 0.31 rad with a maximum flight speed of 0.8 m/s indicated sufficient nimbleness of the individual robot. The detailed results are shown in \figref{fig:ex_fig1} in Supplementary Materials.
\
In the assembled configuration, the vectorable thrust from multiple modules enables three-dimensional morphing and directly contributes to the forces and torques required for aerial manipulation.

To realize an articulated connection between the robot modules, we developed a joint-equipped docking mechanism (\figref{fig:hardware}C), consisting of a pair of conical structures to ensure sufficient interlocking surface. On the male side, there are also a switchable magnetic unit and a retractable stick, which corresponds to a metal plate and a receptor hole on the female side.
When the male and female interfaces contact correctly, the retractable stick automatically inserts into the receptor and completes the mechanical latching (see ``Design'' in Materials and Methods for details).
The load test showed the maximum tensile tolerance of 300 N, reflecting the sufficient connection strength compared with the maximum force of 60 N generated by a single LEGION module.
Each module also has a pair of perpendicular joints that connect the docking structures to the main body and enable the articulated motion in assembled configuration (\figref{fig:hardware}B(iii)).
The pair of conical docking structures allows a radial positional tolerance of 6 cm during approach.
However, the inflight approach is fluctuating due to mutual aerodynamic disturbances. This indicates that mechanical tolerance alone is insufficient to ensure robust docking, motivating the development of our active planning method for inflight docking (see ``Inflight docking planning'' in Materials and Methods).

\clearpage
\subsection*{Inflight docking and morphing}

\begin{figure}
 \centering
 \includegraphics[width=1.0\textwidth]{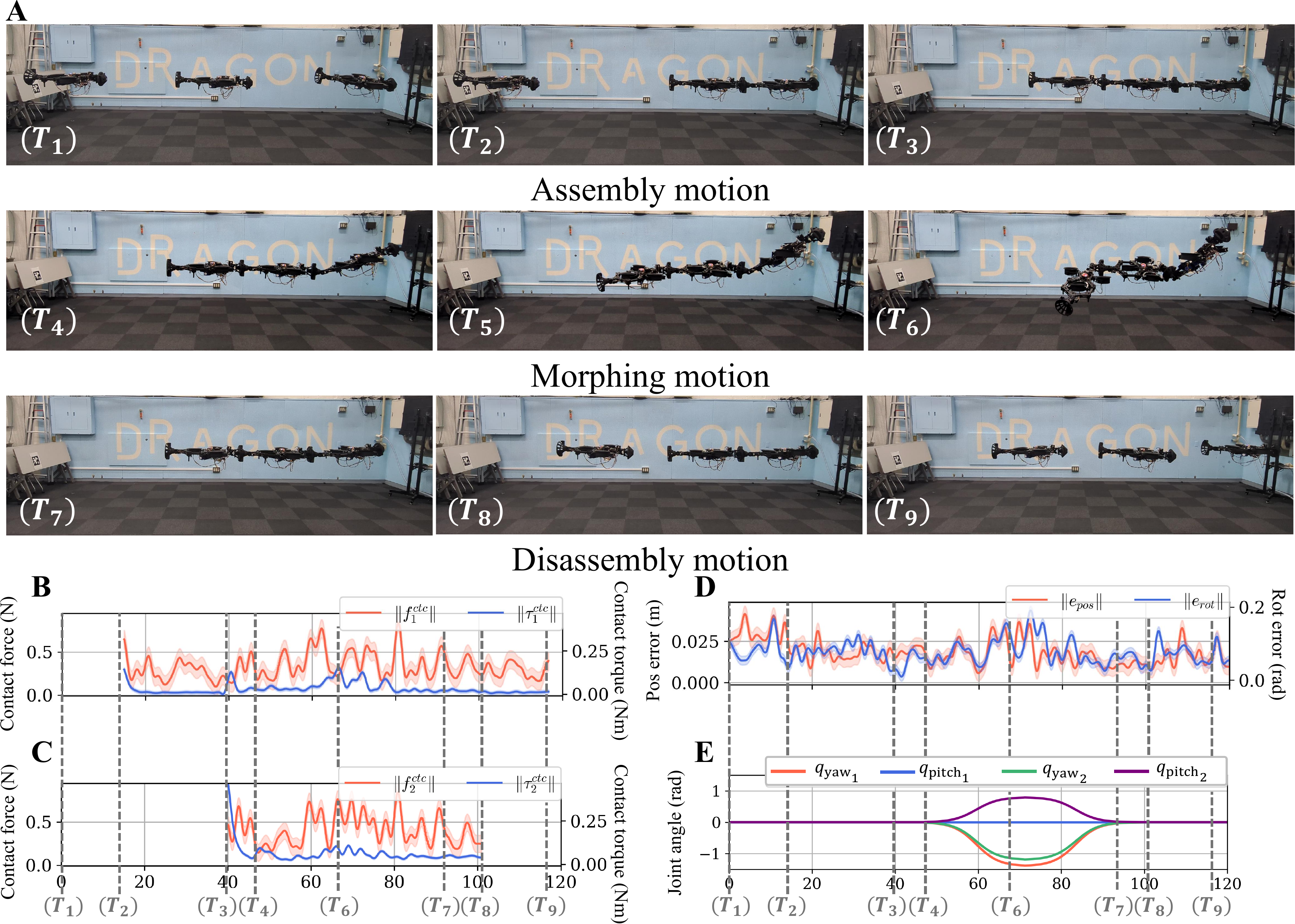}
 \caption{\textbf{Inflight docking and morphing.} 
(\textbf{A}) Three LEGION modules docked in midair ($T_{1} \sim T_{3}$), subsequently performed a morphing motion ($T_{4} \sim T_{6}$), and finally separated again into three individual modules ($T_{7} \sim T_{9}$).
(\textbf{B} and \textbf{C}) Time histories of the contact wrenches between modules. (\textbf{D}) Time histories of the tracking errors of the base module. (\textbf{E}) Time history of the target joint angles. The solid lines in (B-E) represent values processed by a low-pass filter with a cutoff frequency of 5 Hz, and the shaded regions denote the standard deviation over the entire interval.}
 \label{fig:docking_morphing}
\end{figure}

We performed a comprehensive experiment on inflight docking and morphing involving three modules (\figref{fig:docking_morphing}). 
Every robot module autonomously switched from separated mode to assembled mode (\figref{fig:control_framework}).
The experimental results showed that the separated robots could successively assemble to two-module and then three-module configurations (\figref{fig:docking_morphing}A; at time $T_2$ and $T_3$ respectively).
The estimated contact wrenches at two docking points rose abruptly at the moment of docking (the contact torques of 0.13 Nm at time $T_2$ and 0.5 Nm at time $T_3$) due to the approach speed of 0.3 m/s, but rapidly converged to zero which was the desired value.
Subsequently, the assembled modules performed the inflight morphing (\figref{fig:docking_morphing}A; $T_4 \sim T_7$) with a maximum joint speed of  0.12 rad/s.
During morphing, the estimated contact forces fluctuated (\figref{fig:docking_morphing}B, C) because of the unpredictable mutual disturbance; however, the average values were all below 0.3 N (0.03 kg), which is sufficiently smaller than the robot mass of 4 kg.
Moreover, the average values of estimated torques were around 0.05 Nm, and all joint servo loads remained within 0.3 Nm. These results showed the thrust calculated from the contact wrench control contributed to the release of the joint servo load (see ``Contact wrench estimation'' and ``Contact wrench control'' in Materials and Methods).
The average positional and rotational errors of 0.06 m and 0.02 rad showed a comparable flight stability to the individual flight.
Finally, the three modules could be smoothly detached one by one and switched back to the separated mode.
To safely verify the effectiveness of the proposed method, we conducted an ablation study in simulation (see the ``Supplementary Results'' Performance comparison of morphing with and without contact wrench control). In addition, to investigate the scalability of the system, we conducted assembled flight simulations with up to eight modules (see the ``Supplementary Results'' Scalability of multi-connected control framework.).

\begin{figure}
 \centering
 \includegraphics[width=0.85\textwidth]{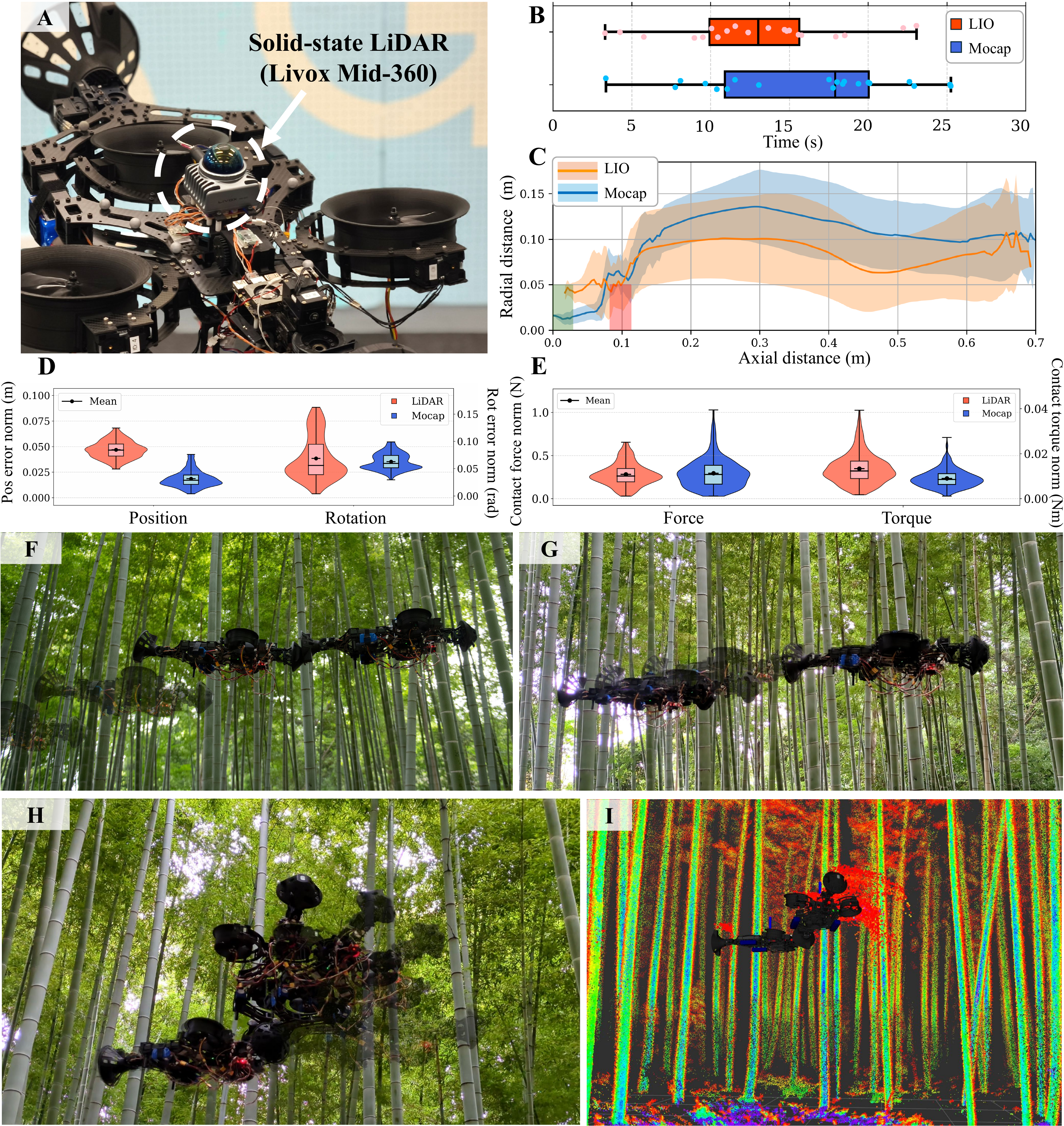}
 \caption{\textbf{Fully autonomous flight using onboard estimation.}
(\textbf{A}) Onboard LiDAR sensor for autonomous localization.
(\textbf{B}) Times of 20 consecutive docking experiments using LiDAR Inertial odometry (LIO) and using an indoor motion-tracking system (Mocap), respectively. 
(\textbf{C}) Relative pose transitions between the pair of male and female docking interfaces. Two solid lines show the averages over 20 trials respectively, and the shaded region represents the standard deviation. The red and green regions indicate the allowable tolerance ranges for transitioning into the ``approach state'' and the ``assembly state'', respectively (see ``Inflight docking planning'' in Materials and Methods); both ranges have a width of 0.025 m and a height of 0.05 m.
(\textbf{D} and \textbf{E}) Flight stability (D) and wrench-control stability (E) in the assembled state.
(\textbf{F}-\textbf{H}) Autonomous self-assembly (F), morphing (G), and self-disassembly (H) in a forest environment. 
(\textbf{I}) Point cloud of the environment constructed by the LiDAR sensor.}
 \label{fig:lidar_flight}
\end{figure}

To further demonstrate the high accuracy and robustness of inflight docking by LEGIONs, we conducted an experiment involving 20 docking trials using only onboard sensors.
The details of onboard sensing for autonomous flight are described in ``Experimental set-up'' of Materials and Methods. As a comparison group, we conducted the same experiment with the assistance of an indoor motion-tracking system.
The results are shown in \figref{fig:lidar_flight}. All 20 trials in both groups succeeded in the inflight docking, and the average time from start to docking was 12.9 s in the case of onboard sensing, which was faster than the average time of 16.0 s by the motion-tracking system (\figref{fig:lidar_flight}B). The slight fluctuation of position estimated by onboard sensors helped the fast shift from alignment state to approach state in the docking planning (\figref{fig:lidar_flight}C and ``Inflight docking planning'' in Materials and Methods). The mechanical docking tolerance of 6 cm was equivalent to allowing up to 3 cm of position-estimation error per module, which comfortably accommodated the maximum localization error of 2.8 cm arising from onboard sensing. 
The state-estimation error also did not affect the docking success decision in most cases since the relative-position error threshold for successful docking was set to 5 cm. In two out of twenty trials, however, the estimated relative error did not satisfy this threshold despite the modules being physically docked. In these cases, as described in the ``Inflight docking planning'' in Materials and Methods, since the inter-module contact wrench exceeded the allowable wrench threshold, the docking was judged successful. These results showed the sufficient robustness of our docking planning against the sensing uncertainty and the disturbance.
We also evaluated the stability in the assembled mode.
Compared with the average positional tracking error of 0.02 m in the motion-tracking system, the tracking error of 0.04 m with only onboard sensing was larger but still sufficient to maintain stable assembly (\figref{fig:lidar_flight}D).
Notably, the estimation of contact wrench was not significantly different between the two cases (\figref{fig:lidar_flight}E).
These results showed that the proposed contact wrench control remained effective even when position estimation contains uncertainty and bias.

To evaluate the autonomous capability of the LEGION system in real environments,  a self-reconfiguration experiment was also conducted in a GNSS-denied forest environment (\figref{fig:lidar_flight}F-H).
The docking time was 12.2 s, and the positional and rotational tracking errors during morphing were 0.09 m and 0.06 rad, respectively.
This experiment demonstrates that LEGION can perform assembly, disassembly, and morphing outdoors in a manner comparable to indoor operation.

\clearpage
\subsection*{Cooperative aerial manipulation}

\begin{figure}
 \centering
 \includegraphics[width=\textwidth]{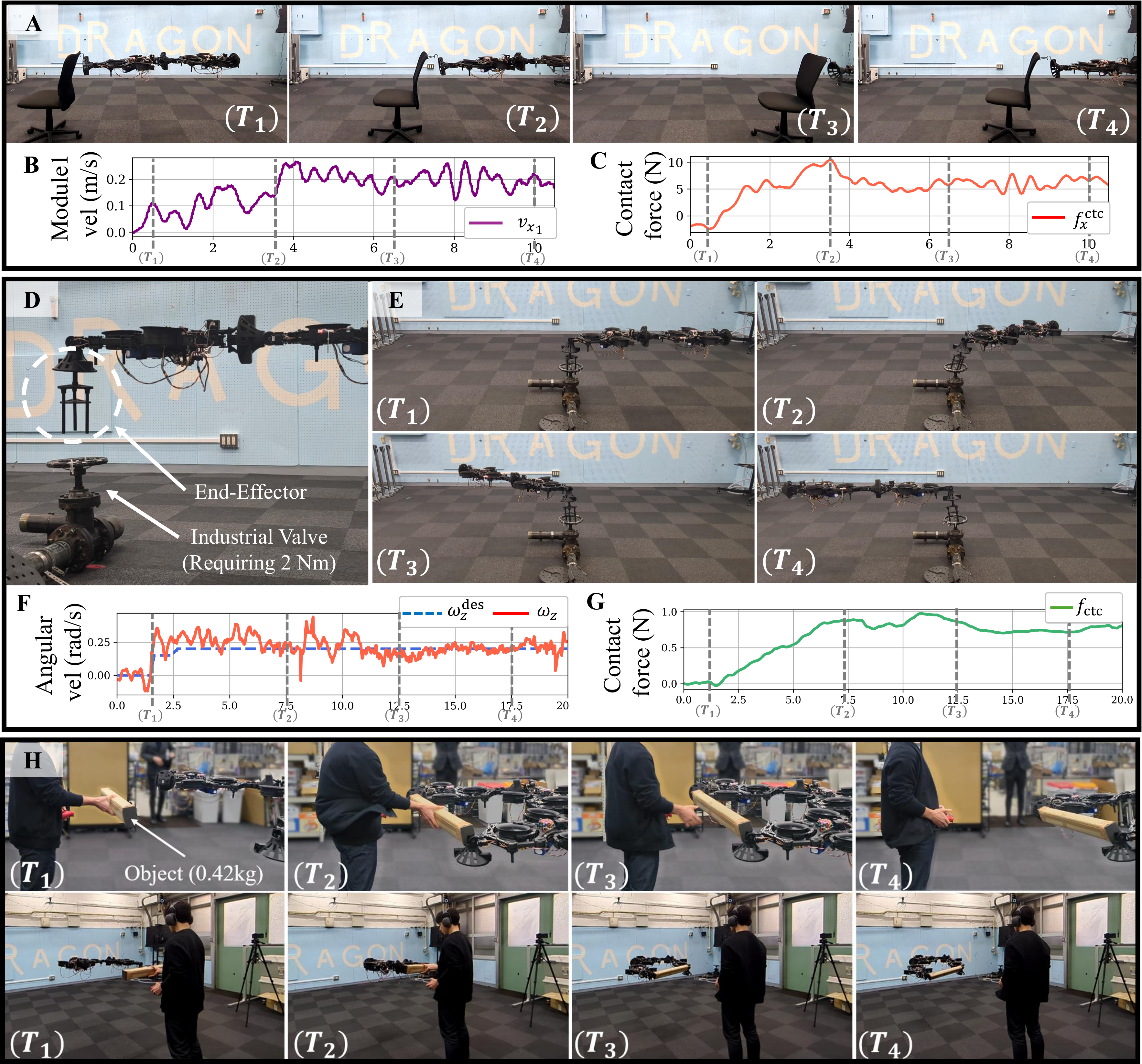}
 \caption{\textbf{Manipulation tasks in different assembled configurations.}
 (\textbf{A}) Object-pulling task performed in the two-module assembled configuration.
(\textbf{B} and \textbf{C}) Time histories of the base module velocity (B) and the contact wrench applied to the object (C). 
(\textbf{D} and \textbf{E}) Valve-rotation task performed in the two-module assembled configuration. Experimental setup (D) and snapshots (E).
(\textbf{F} and \textbf{G}) Time histories of the base module angular velocity (F) and the contact wrench between modules (G).
(\textbf{H}) Object-grasping task performed in the three-module assembled configuration. }
 \label{fig:manipulation_results}
\end{figure}

To validate LEGION's cooperative manipulation capacity, we first conducted experiments with four manipulation motions: pushing, pulling, rotating, and grasping (\figref{fig:manipulation_results}).
We commanded the two assembled modules to push or pull a chair with a reference moving speed of 0.2 m/s. Both modules gradually converged to this speed, while the estimated contact force corresponding to the friction between the chair and the floor increased to 5 N (\figref{fig:manipulation_results}A-C).
We then evaluated the performance of rotating an industrial valve. To manipulate the valve, a branched gripper was attached to the end of the inner module (\figref{fig:manipulation_results}D).
During rotating, the outer module was required to appropriately assist the inner one by generating appropriate operation force. 
As shown in \figref{fig:manipulation_results}E-G, a stable rotating speed around a reference 0.2 rad/s was achieved by properly controlling the desired contact force at the docking point along the circumferential direction (see ``Contact wrench control'' in Materials and Methods). 

\begin{figure}
 \centering
 \includegraphics[width=1.0\textwidth]{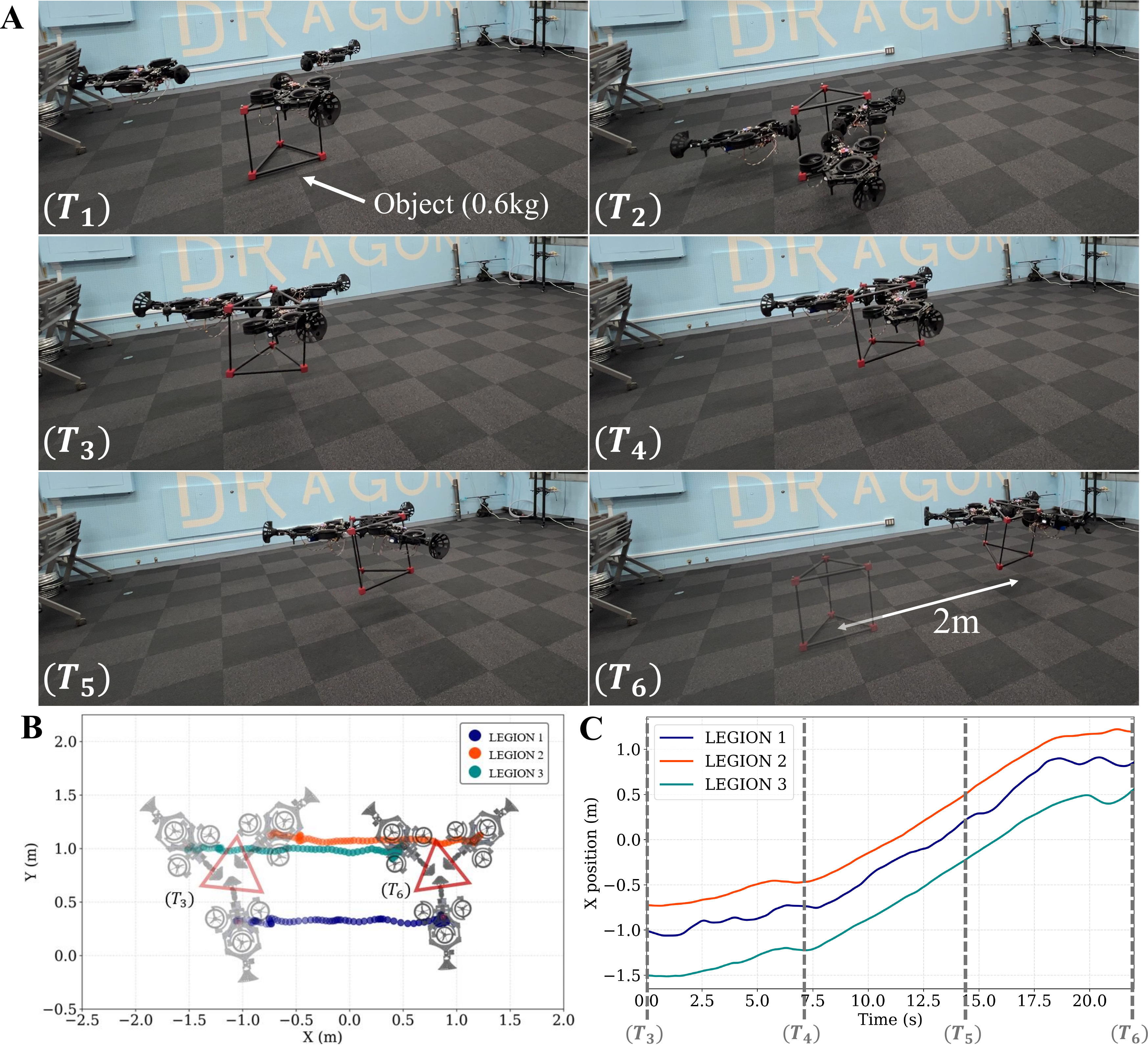}
 \caption{\textbf{Swarm transportation.}
(\textbf{A}) Three LEGION modules transported a large object with 0.5--m edges in a synchronized manner while maintaining a triangular formation.
(\textbf{B} and \textbf{C}) Trajectories of each module in the $xy$-plane during the task (B), and time histories of their positions along the $x$-axis (C).
}
 \label{fig:swarm_manipulation}
\end{figure}

To demonstrate a manipulation capability that leverages joint motion, an object grasping experiment was conducted. In this experiment,  three-module aggregation grasped a 0.42 kg stick-like object from a human during flight and subsequently transported it (\figref{fig:manipulation_results}H). The object was grasped using the two ends of the chain as end effectors. In this case, no explicit control of the contact forces acting on the object was applied; instead, it was sufficient to set the target joint angles such that the end effectors penetrated 5 cm into the object.
For a larger object, the separated mode where the robots individually select the valid contact points can ensure a better transportation performance than the assembled mode. We then conducted an experiment where a swarm of three LEGION modules cooperatively but separately transported a triangular prism with a side length of 0.5 m and a mass of 0.6 kg (\figref{fig:swarm_manipulation}). Three modules carried each edge of the object, and moved synchronously at a speed of 0.09 m/s. Although some fluctuating motion caused by mutual aerodynamic interference at close range could be observed across all modules, the object was successfully moved with a certain distance of 2 m owing to the multiple contact points by these modules.

These five tasks can be considered as the primitives in aerial manipulation, and the flight stability of the assembled configuration during manipulation was comparable to that of single-module flight.
These demonstrations highlight the potential of the proposed cooperative manipulation in more demanding scenarios and further expand the capabilities of swarm aerial robots.

\section*{Discussion}
This study proposes an aerial modular robot named LEGION, inspired by self-reconfigurable biological systems such as fire ants. LEGION modules can dynamically realize various body configurations by assembling and disassembling in midair, thereby altering task characteristics such as body size, shape, and degrees of freedom of joints.
Through this capability, LEGION bridges the characteristics of both small and large aerial robots, significantly enhancing the versatility of aerial robotics.
\subsection*{LEGION's position as an aerial robotic system}
\begin{figure}
 \centering
 \includegraphics[width=1.0\textwidth]{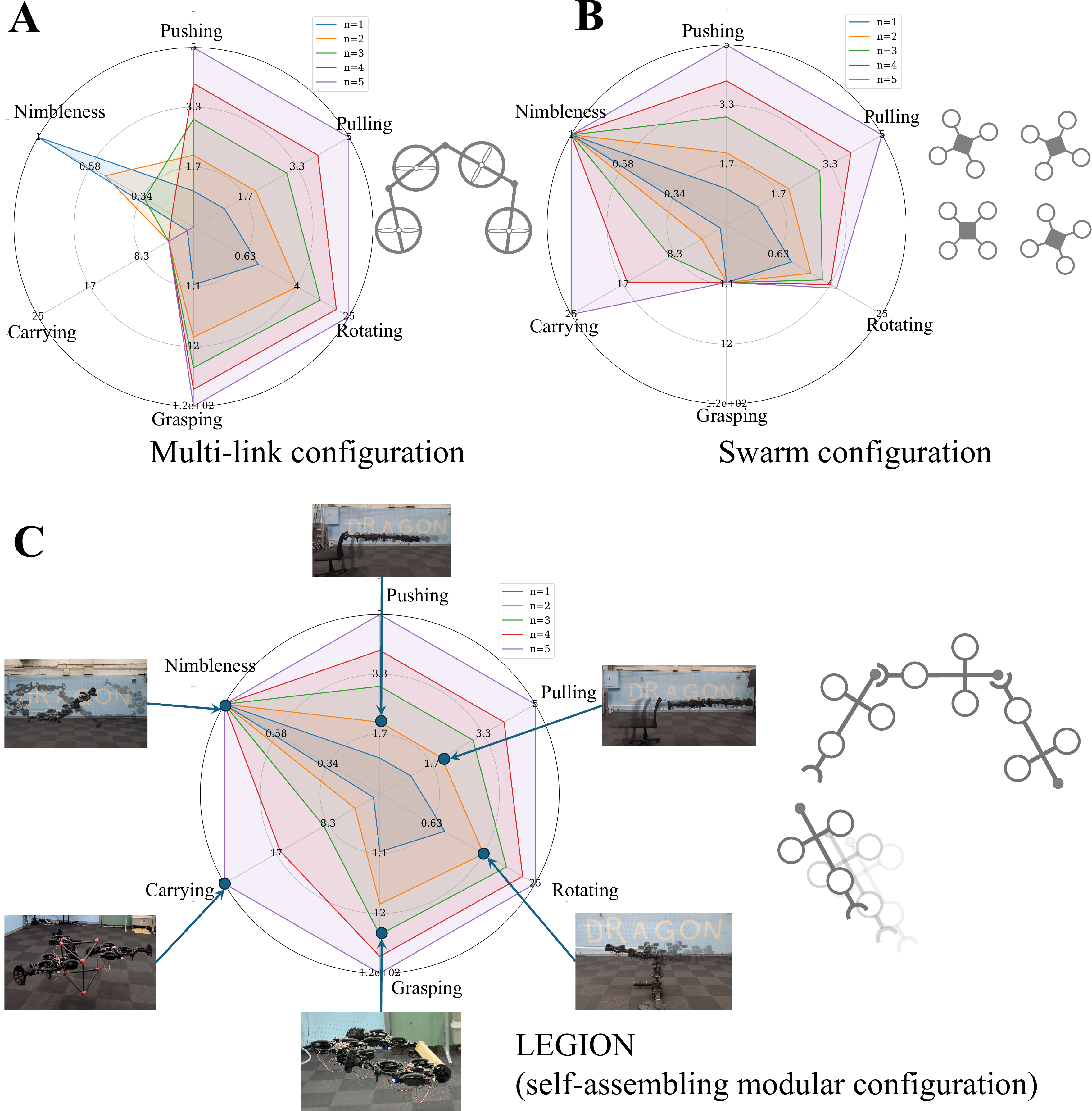}
 \caption{\textbf{Manipulability and Nimbleness of different types of aerial robot configurations}. 
(\textbf{A}) The overall performance of the multi-link configuration.
(\textbf{B}) The overall performance of the swarm configuration.
(\textbf{C}) The overall performance of modular aerial robot LEGION. Here, $n$ denotes the number of links in the multi-link configuration, the number of agents in the swarm configuration, and the number of LEGION modules (for details on how each axis metric is computed, see the ``Supplementary Methods'' Task metrics). In \textbf{c}, the blue points and corresponding figures show the task experiments conducted in this study.}
\label{fig:manipulation_primitives}
\end{figure}

To demonstrate that inflight self-organization is an effective strategy for achieving the versatility in aerial manipulation, we first quantified the five manipulation primitives---pushing, pulling, rotating, grasping, and carrying (see the ``Supplementary Methods'' Task metrics).
We then evaluated whether the proposed modularity could realize all these manipulation primitives more effectively than existing approaches. As representative baselines, we selected both the multi-link configuration and the swarm configuration (\figref{fig:manipulation_primitives}).
Here, the number of modules in LEGION, the number of links in multi-link configuration, and the number of units in swarm configuration were treated as scaling parameters. We examined how performance in each primitive varied with respect to these scaling parameters. 
The multi-link configuration can transmit internal forces between links during morphing; thus, its performance in pushing, pulling, rotating, and  grasping, improves effectively as the number of links increases (\figref{fig:manipulation_primitives}A). However, because the number of end-effectors remains limited to two with an increased number of links, stability in object support does not scale, meaning that performance in carrying cannot be enhanced by adding links. Moreover, as the number of links increases, the system becomes less nimble, making operation in confined spaces increasingly difficult.
Conversely, because each unit in the swarm configuration can contribute an independent support point, their payload-carrying capability increases linearly without the loss of nimbleness. However, the swarm configuration cannot transmit internal forces between units. Thus, they are fundamentally incapable of grasping, and their scaling efficiency in rotating is significantly lower than that of the multi-link configuration (\figref{fig:manipulation_primitives}B).
The LEGION system can operate as a swarm in the separated mode and as a multi-link configuration in the assembled mode, thereby combining the complementary strengths of the two approaches. This enables the system to scale various manipulation capabilities effectively as the number of modules increases without compromising maneuverability(\figref{fig:manipulation_primitives}C).
Although the concept of self-assembling aerial robot is shared with several prior studies \cite{dockable-multirotor-2025-Access, trady-2023-AIS, beatle-2024-TMech}, these works primarily focused on the assembly and disassembly motion itself, rather than on expanding and validating task performance through variations in body size, shape, and degrees of freedom of joints. In this respect, the design philosophy of LEGION is fundamentally distinct. The task versatility of LEGION is clearly characterized by its articulated module design and by the distributed assembled flight control framework based on contact wrench control, which enables joint motion even in the assembled state. 
\subsection*{Limitations and future works}
While LEGION's exceptional self-reconfiguration capability enables remarkable adaptability to a wide range of tasks, each module must include additional components such as docking and joint mechanisms, which compromises energy efficiency when operating individually. 
This issue is not unique to LEGION but represents a general dilemma shared by conventional self-assembling aerial robots and multi-link aerial robots. 
Although energy efficiency optimization was not a primary focus of this study, we evaluated the ratio of docking mechanism mass to total mass ($D_{\text{mass}}$) and flight duration and compared these parameters with other aerial robot systems (see \tabref{tab:a-msrr} in the Supplementary Materials).
The $D_{\text{mass}}$ is defined as $D_{\text{mass}}:=\frac{m_{\text{dock}}}{m_{\text{total}}}$. 
LEGION exhibited a $D_{\text{mass}}$ of 0.12 and a continuous flight time of 6 minutes. 
These specifications are comparable to the average values of state-of-the-art self-assembling aerial robots and multi-link aerial robots, suggesting that LEGION's design is appropriate. 
For detailed comparisons of energy efficiency between LEGION and other systems, refer to the ``Supplementary Results'' Hardware efficiency analysis.
When prioritizing energy efficiency, it can be improved by trading off other performance factors. 
For instance, reducing the yaw joint (see \figref{fig:hardware}C(i)) range by 20 degrees allows the rotor diameter to increase from 5 inches to 6 inches, which ideally enhances thrust by about 70\%. 
Alternatively, reducing the thrust vectoring DoF (see \figref{fig:hardware}B(ii)) from two to one can reduce the number of actuators and achieve weight reduction. 
This configuration can achieve high efficiency when the body attitude remains near horizontal, but when the attitude tilts significantly, thrust decreases and efficiency is reduced.
Achieving both high energy efficiency and task capability beyond such trade-offs remains a major challenge in aerial robotics and will be addressed in future work.

Regarding the inter-module communication, the wireless communication-based architecture used in this study performed sufficiently well under the tested conditions; however, for larger-scale missions or high-noise environments, wired connections for inter-module communication and power systems may improve LEGION's robustness. 
Furthermore, high-throughput wired communication could allow high-frequency acquisition of each module's state, potentially enabling centralized model-based optimal control for the entire assembled structure. 
The optimal design of inter-module networks and the development of flexible centralized control strategies are important directions for future research.

In summary, we proposed an articulated aerial modular robot system capable of self-assembly and reconfigurable to overcome the limited task versatility of conventional aerial robots caused by hardware constraints. 
The proposed system enables a single platform to execute diverse tasks, including complex trajectory tracking and high-load manipulation. 
We believe that this platform will open new possibilities for aerial robot applications in environments requiring multifunctional capabilities, such as industrial plants and disaster sites.
\section*{Materials and Methods}
\subsection*{Design}
\subsubsection*{Rotor configuration}
Regarding the flight performance of individual modules, each LEGION must be fully-actuated to enable the precise alignment between modules during docking. 
In general, there are two possible methods to achieve full actuation in multirotor aerial robots: 
(1) using six or more rotors fixed at tilted angles, or (2) using three or more rotors capable of thrust vectoring. 
Since the fixed-tilt configuration easily causes inter-rotor aerodynamic interference and thus thrust loss, we adopt rotors that can individually vector their thrust in two degrees-of-freedom (dof), allowing arbitrary thrust orientation (\figref{fig:hardware}B(ii)).
Rotor vectoring is achieved by a pair of perpendicular servo motors.
Then, each vectorable rotor can generate three-dimensional force, which is equal to three control inputs. We deploy three rotors to generate nine control inputs to handle six dof of full pose motion, ensuring the robustness due to the control redundancy. 
Ducts are used to enclose all three rotors, providing higher thrust efficiency and protection from collisions. These ducts are fabricated via 3D printing with Polyamide 6-Carbon Fiber Composite (PA6-CF).
Thrust testing confirmed that the duct improved thrust by approximately 20\% compared to the non-ducted configuration.
Each rotor with 5 inch polycarbonate tri-blade propeller can generate a maximum thrust of 21 N, and thus the maximum lifting force generated by three rotors is 63 N.

\subsubsection*{Docking mechanism}
The docking system of LEGION combines a mechanical locking mechanism with a magnetic coupling structure.
The mechanical locking mechanism (see \figref{fig:ex_fig4}C in the Supplementary Materials) is inspired by the retractable mechanism of a retractable pen.
When a pair of docking interfaces becomes sufficiently close, the stick of the male interface is ``clicked'' by the drogue of the female interface, and ejected into the corresponding receptor in the female interface. This passive locking process allows the connection to be completed precisely at the instant of docking, eliminating the need to actively maintain alignment under inflight disturbances.
For detaching, the stick is retracted into the initial idling position via a wire driven by a spool and a small servo (see \figref{fig:ex_fig4}B in the Supplementary Materials).
The magnetic coupling structure is composed of a magnetic switching mechanism \cite{switchable-permanent-magnet-patent} embedded at the top of male interface, and a metal plate at the bottom of the female interface.
The magnetic tensile force, which has a  maximum of 265 N, is switched by a knob rotated by the same servo that winds the wire for the retractable stick. The wire winding and magnetic knob rotation are synchronized to switch between attaching and detaching. For the detailed operation process of docking mechanism, please refer to \figref{fig:ex_fig4} in the Supplementary Materials.

\subsubsection*{Joint configuration}

A pair of orthogonal joints are placed at the  ends of the main body, which offer the pitch rotation for the female docking interface and yaw rotation for the male docking interface, respectively. 
This configuration enables a three-dimensional articulated motion in assembled mode by the joint angles $\boldsymbol{q} \in \mathbb{R}^{2n}$.
Each joint is actuated by a compact servo which can generate a maximum torque around 2.5 Nm. We also use a pair of pulleys (the radius ratio is 1:2) to double the output torque to 5 Nm, but the maximum joint rotation speed decreases to 2.6 rad/s.

\subsubsection*{Airframe}
The frame of LEGION is primarily constructed from Carbon Fiber Reinforced Plastic (CFRP) and PA6-CF, both of which offer lightweight and high stiffness. 
The whole weight of the main body is 3.9 kg, and the total flight time is 6 min achieved by  a pair of 6S LiPo batteries with the total capacity of 6000 mAh. Detailed specifications of the robot module are summarized in \tabref{tab:bom}.

\subsubsection*{Electronics}
All computation in LEGION is executed in an onboard compact computer (Khadas VIM4) and a flight control board (STM32H7). These boards are all powered by the same batteries for rotors and servos.
Communication between the computer and the flight control board is handled via UART, while the commands from the flight control board to the electronic speed controllers are sent using the DShot protocol.
The flight control board also sends the target angles to the servos via TTL and RS485, and the servos execute the position control to track the desired angle. 
At the same time, the servos send back the measured joint angles and torques to the flight control board and then to the onboard computer to support following flight control and assembly control.

\subsection*{Flight control}
The dynamical model of a single LEGION is:
\begin{align}
 m\ddot{\boldsymbol{r}}_c &= \boldsymbol{R}_c\boldsymbol{f}_c+m\boldsymbol{g}, \\
 \boldsymbol{I}\dot{\boldsymbol{\omega}}_c &= \boldsymbol{\tau}_c - \boldsymbol{\omega}_c\times\boldsymbol{I}\boldsymbol{\omega}_c,
 \label{eq:motion_eq}
\end{align}
where $m$ denotes the total mass of a single module, $\boldsymbol{g}$ is the gravitational acceleration, $\boldsymbol{I}$ is the inertia tensor of the body. 
$\boldsymbol{r}_c$ and $\boldsymbol{R}_c$ are the centroidal position and orientation, respectively.
$\boldsymbol{f}_c$ and $\boldsymbol{\tau}_c$ represent the total centroidal force and torque which are defined as 
\begin{equation}\label{equation:allocation}
 \begin{bmatrix}
  \boldsymbol{f}_c \\
  \boldsymbol{\tau}_c
  \end{bmatrix}
 =
  \boldsymbol{Q}
 \begin{bmatrix}
     \boldsymbol{\lambda}_{1} \\ \boldsymbol{\lambda}_{2} \\ \boldsymbol{\lambda}_{3}
 \end{bmatrix},
\end{equation}
where $\boldsymbol{Q}$ denotes the allocation matrix determined by the geometric configuration of each rotor, and  $ \boldsymbol{\lambda}_{i} = \begin{bmatrix}\lambda_{i,x}&\lambda_{i,y}&\lambda_{i,z}\end{bmatrix}^{\mathsf{T}}$ represents the three-dimensional force vector generated by each rotor. 

We then use proportional-integral-derivative control to achieve the desired centroidal force and torque:
\begin{align}
  \label{eq:pid_f}
  \boldsymbol{f}^{\text{des}}_c
  &= m\boldsymbol{R}_{c}^{-1}
  \left(
    k_{f,\text{p}}\boldsymbol{e}_{r}
    + k_{f,\text{i}}\!\int\!\boldsymbol{e}_{r}\,dt
    + k_{f,\text{d}}\dot{\boldsymbol{e}}_{r}
  \right) - m\boldsymbol{g}, \\
  \label{eq:pid_t}
 \boldsymbol{\tau}^{\text{des}}_c
  &=
  \!\boldsymbol{I}
  \left(
    k_{\tau,\text{p}}\boldsymbol{e}_{R} 
    + k_{\tau,\text{i}}\!\int\!\boldsymbol{e}_{R}\,dt
    + k_{\tau,\text{d}}\dot{\boldsymbol{e}_{R}}
  \right)
  + \!\boldsymbol{\omega}_c \times \!\boldsymbol{I}\,\!\boldsymbol{\omega}_c,
\end{align}
where $\boldsymbol{e}_r$ and $\boldsymbol{e}_{R}$ denote the tracking error between the desired and actual pose \cite{se3-control-2010-CDC}.
Each $k$ represents the gain of this closed-loop control.

The desired thrust vector can then be calculated by minimizing the thrust effort as follows:
\begin{align}
\label{eq:control_allocation}
 \begin{bmatrix}
     \boldsymbol{\lambda}^{\text{des}}_{1} \\ \boldsymbol{\lambda}^{\text{des}}_{2} \\ \boldsymbol{\lambda}^{\text{des}}_{3}
 \end{bmatrix}
 =     \boldsymbol{Q}^{\#}\left( \begin{bmatrix} \boldsymbol{f}^{\text{des}}_c \\ \boldsymbol{\tau}^{\text{des}}_c \end{bmatrix} + \mathbf{w}^{\text{ff}}\right),
\end{align}
where pseudo-inverse $\boldsymbol{Q}^{\#}$ can be computed using the Moore-Penrose method \cite{pseudo-inverse}.
$\mathbf{w}^{\text{ff}}$ denotes the feed-forward term which can compensate the disturbance. In assembled mode and cooperative manipulation, we use this term to control the contact wrench between the interlocked modules.

The magnitude of the thrust for each rotor can be finally obtained as $T_{i} = \|\boldsymbol{\lambda}_{i}\|$, while the target vectoring angles are calculated as $\phi_i = \text{tan}^{-1}(-\tfrac{\lambda_{i,y}}{\lambda_{i,z}})$ and $\theta^{\text{des}}_i = \text{tan}^{-1}(\tfrac{\lambda_{i,x}}{-\lambda_{i,y} \text{sin}(\phi_{i}) + \lambda_{i,z} \text{cos}(\phi_{i})})$.
This control flow can minimize the nonlinear complexity caused by the trigonometric functions of vectorable rotor, and enables a fast computation in our flight control board at a frequency of 500 Hz, which is more efficient compared with computationally expensive optimization process such as the nonlinear model predictive control \cite{beetle-omni-2024-RAL}.

\subsection*{Inflight docking planning}

\subsubsection*{State machine}

The operation procedure of the docking is represented as a simple state machine with three states: alignment, approach, and assembly.
In the alignment state, the follower module aligns itself to a line related to the pose of the leader module (\figref{fig:control_framework}). Once the alignment converges sufficiently, the follower module shifts to the approach state and moves with a constant speed toward the leader. This approach momentum ensures a click on the top of the retractable stick and then triggers the docking mechanism to interlock two modules. If the relative radial position error during approaching exceeds the docking tolerance of 5 cm, a recovery action will execute and the follower module will immediately return to the alignment state.
The further transition to the assembly state is determined by considering both the relative position of the two docking modules and the accumulated contact wrench at docking interface. If pose estimates are accurate, only the former condition suffices.
Otherwise, the latter condition is used: as the two modules continue to generate velocity toward the docking direction, a contact wrench accumulates at the docking interface. Once this wrench exceeds a threshold, the state machine declares successful assembly and transitions to the assembly state.
The contact force can be estimated by \equref{eq:wrench_observer}.
The reason for using this dual-condition structure is to increase robustness against the position and force estimation uncertainties---especially in outdoor environments where all estimations rely solely on onboard sensors. 
Because the mutual disturbance between laterally approaching aerial robots is more difficult to model than the vertical-stack flight\cite{flyingtoolbox-2025-Nature, aerial-charging-2025-ICRA}, this model-free state machine with recovery function is more reliable than the trajectory tracking control which requires an accurate model for planning \cite{fast-landing-2025-JFR}.

\subsubsection*{External wrench estimation}
The external force and torque--the external wrench--are estimated with the robot states based on the momentum based observer\cite{momentum-based-observer-2025-ICRA}:
\begin{equation}
\hat{\mathbf{w}}^{\mathrm{ext}}(t) =K_I\!\left[\, \boldsymbol{p}(t)-\boldsymbol{p}(t_0) -\!\!\int_{t_0}^{t}\!\! \Big(J^{\top}\!\mathbf{w}_{\mathrm{known}}(s)
+\hat{\mathbf{w}}^{\mathrm{ext}}(s)-\boldsymbol{n}\Big)\,ds\right],
\label{eq:wrench_observer}
\end{equation}
where $\mathbf{w}_{\mathrm{known}} =\boldsymbol{Q}\begin{bmatrix}\boldsymbol{\lambda}_1&\boldsymbol{\lambda}_2&\boldsymbol{\lambda}_3\end{bmatrix}^{\mathsf{T}} +\boldsymbol{B}\boldsymbol{\tau}$ denotes the active wrench generated by the rotors and the joints. $\boldsymbol{\tau} = [\tau_\mathrm{yaw}\ \tau_\mathrm{pitch}]^{\mathsf{T}}$ is the joint torque vector, which is transformed into the wrench vector by $\boldsymbol{B}\in\mathbb{R}^{6\times2}$. $\boldsymbol{n}=\begin{bmatrix} m\boldsymbol{g}& \boldsymbol{\omega}\times\boldsymbol{I}\boldsymbol{\omega}\end{bmatrix}^{\mathsf{T}}$ is the inertial term of a single module.
$\boldsymbol{K}_I\in \mathbb{R}^{6\times6}$ and $\boldsymbol{J}\in \mathbb{R}^{6\times6}$ denote the observer gain matrix and transformation matrix, respectively. 
$\boldsymbol{p}$ is the generalized momentum which can be calculated from the estimated linear and angular velocities. This momentum-based observer does not need the acceleration, and thus can avoid the high-frequency noise from the IMU measurement.
$t_0$ denotes the initial time when the robot begins hovering.
The external wrench estimated  in each module $\hat{\mathbf{w}}^{\mathrm{ext}}_{i}$ is shared among other modules for subsequent contact wrench estimation.

\subsection*{Contact wrench estimation}

The external wrench of the $i$-th module $\mathbf{w}^{\mathrm{ext}}_i$ can be further decomposed as follows:
\begin{equation}
\mathbf{w}^{\mathrm{ext}}_i
=  \mathbf{w}^{\mathrm{ctc}}_{i-1} 
- \mathbf{w}^{\mathrm{ctc}}_{i} 
+ \mathbf{w}^{\mathrm{dist}},
\label{eq:connected_dynamics}
\end{equation}
where $\mathbf{w}^{\mathrm{ctc}}_{i}$ and $\mathbf{w}^{\mathrm{ctc}}_{i-1}$ denotes the contact wrenches at the both sides of docking interface (\figref{fig:control_framework}), and $\mathbf{w}^{\mathrm{dist}}$ represents the common disturbance (e.g., wind disturbance) and some other constant force (e.g., the mass of a grasped object) among all connected modules.

Similar to the coupled vibration of multi-mass systems \cite{coupled-vibration}, the contact wrench $\mathbf{w}^{\mathrm{ctc}}_{i}$ in connected modules can be also written in a linear system:
\begin{equation}
 A\begin{bmatrix}\mathbf{w}^{\mathrm{ctc}}_{1} \\ \vdots \\ \mathbf{w}^{\mathrm{ctc}}_{N}   \end{bmatrix}  = \begin{bmatrix}\boldsymbol{d}_1\\\vdots\\\boldsymbol{d}_n\end{bmatrix},
\label{eq:sim_eq_ctc}
\end{equation}
where $A$ is a band matrix composed of the geometric information of the chained configuration, and $\boldsymbol{d}_{i}=\mathbf{w}^{\mathrm{ext}}_i - \mathbf{w}^{\mathrm{dist}}$.
The common disturbance wrench $\mathbf{w}^{\mathrm{dist}}$ can be calculated by averaging the estimated external wrenches:
$
\mathbf{w}^{\mathrm{dist}}
=\frac{1}{N}\sum_{k=1}^{N}
\hat{\mathbf{w}}^{\mathrm{ext}}_k
$.

This linear system has infinite solutions for $\mathbf{w}^{\mathrm{ctc}}_{i}$; however, the contact wrench at the end module satisfies $\mathbf{w}^{\mathrm{ctc}}_N = \mathbf{0}$, which offers the boundary condition for this linear system. Therefore, it is possible to find a unique solution for the chain structure by solving following optimization problem:
\begin{equation}
 \min_{\mathbf{w}^{\text{ctc}}}\ \tfrac12\|A\mathbf{w}^{\text{ctc}}-\boldsymbol{d}\|_2^2\ \ \text{s.t.}\ \ \boldsymbol{C}_{\text{ctc}} \mathbf{w}^{\text{ctc}}=\boldsymbol{0}
\label{eq:ctc_min_problem}
\end{equation}
where $\mathbf{w}^{\text{ctc}} = \big[\mathbf{w}^{\text{ctc}}_{1}\; \hdots\; \mathbf{w}^{\text{ctc}}_{N}]$, $\boldsymbol{d} := \big[\boldsymbol{d}_{1}\; \hdots\; \boldsymbol{d}_{N}]$, and $\boldsymbol{C}_{\text{ctc}} := \big[\,0_{6\times 6(N-1)}\ \ I_6\,\big]$ respectively.

\subsection*{Contact wrench control}

To obtain the feed-forward term for $i$-th module $\mathbf{w}^{\mathrm{ff}}_{i}(t+1)$ in \equref{eq:control_allocation} to generate the desired contact wrench $\bar{\mathbf{w}}^{\mathrm{ctc}}_i$, we consider the change in $\mathbf{w}^{\mathrm{ctc}}_i$ between discrete steps $t$ and $t{+}1$:
\begin{equation}
\bar{\mathbf{w}}^{\mathrm{ctc}}_i(t+1)
= 
\mathbf{w}^{\mathrm{ctc}}_i(t) + \mathbf{w}^{\mathrm{ff}}_{i}(t+1) - \mathbf{w}^{\mathrm{ff}}_{i+1}(t{+}1),
\label{eq:ff_dynamics}
\end{equation}
where $\bar{\mathbf{w}}^{\mathrm{ctc}}_i(t+1)$ is given as the desired contact wrench at the $i$-th docking point.

Similar to \equref{eq:sim_eq_ctc}, \equref{eq:ff_dynamics} can be also summarized as a linear system with respect to $\mathbf{w}^{\mathrm{ff}}_{i}(t+1)$:
\begin{equation}
 H\begin{bmatrix}\mathbf{w}^{\mathrm{ff}}_{1}(t) \\ \vdots \\ \mathbf{w}^{\mathrm{ff}}_{N}(t)   \end{bmatrix}  = \begin{bmatrix}\boldsymbol{\delta}_1(t)\\\vdots\\\boldsymbol{\delta}_n(t)\end{bmatrix},
\label{eq:sim_eq_ff}
\end{equation}
where $H$ is another band matrix composed of the geometric information of the chained configuration, and $\boldsymbol{\delta}_i(t) = \bar{\mathbf{w}}^{\mathrm{ctc}}_i(t+1) - \mathbf{w}^{\mathrm{ctc}}_i(t)$.

In assembled mode, the base module is designed as a leader and thus does not perform contact wrench control. Other modules are required to perform both pose tracking and contact wrench control. Therefore, the constraint $\mathbf{w}^{\text{ff}}_{\text{base}}=\boldsymbol{0}$ is imposed, and \equref{eq:sim_eq_ff} has a unique solution, obtained in the same manner as for \equref{eq:ctc_min_problem}.
Given that communication delays and observation noise are inevitable in real environments, the final feed-forward term for \equref{eq:control_allocation} needs following smoothing process with a coefficient $\gamma \in (0,1]$:
\begin{equation}
\mathbf{w}^{\mathrm{ff*}}_{i}(t+1) =  \gamma \mathbf{w}^{\mathrm{ff}}_{i}(t+1) + (1-\gamma) \mathbf{w}^{\mathrm{ff*}}_{i}(t).
\label{eq:ff_smoothing}
\end{equation}
A detailed discussion of the convergence properties of the proposed control flow is provided in the ``Supplementary Methods'' Convergence of the control flow for multi-connected flight.

The desired contact wrench is designed according to the tasks. For example, in the motion of pushing and pulling of a chair (\figref{fig:manipulation_results}A), all desired contact wrenches are set as zero ($\bar{\mathbf{w}}^{\mathrm{ctc}}_i=\boldsymbol{0}$), because all modules are assumed to generate equal pushing or pulling force. 
For the valve rotation task (\figref{fig:manipulation_results}E), the resistance torque from the valve affected the rotation speed. Therefore we design a closed-loop force control to ensure the constant rotation speed:
\begin{equation}
f^{\text{des}}_{\text{ctc}} = k_{\text{p}} (\omega_z^{\text{des}} - \omega_z) + k_{\text{i}}\!\int\!(\omega_z^{\text{des}} - \omega_z)\,dt,
\label{eq:valve_pi_control}
\end{equation}
where $f^{\text{des}}_{\text{ctc}}$ is the second element of $\bar{\mathbf{w}}^{\mathrm{ctc}}_1$ in this task, and other elements are all zero.

\subsection*{Trajectory generation}

Since LEGION has the omni-directional maneuvering ability, we can design the flight trajectory based on a cubic spline to ensure independent smoothness regarding the position and orientation respectively (\figref{fig:ex_fig1}B). The target pose and twist at each time are substituted into the feed-back control of \equref{eq:pid_f} and \equref{eq:pid_t}, while the target acceleration is considered as a feedforward term $\mathbf{w}^{\text{ff}}$ in \equref{eq:control_allocation}.

For the morphing in assembled mode, we first design the trajectory for the whole chained structure which can be composed to the motion of base module and the joint angles of the chain: $\boldsymbol{x} = \begin{bmatrix}\boldsymbol{r}_{\text{base}}&\boldsymbol{\alpha}_{\text{base}}&\boldsymbol{q}\end{bmatrix}$, where $\boldsymbol{\alpha}_{\text{base}}$ is Euler angle vector. Given the joint motion, it is necessary to ensure the smoothness on jerk (the derivative of acceleration). Then, a polynomial of the fifth degree is designed as:
\begin{equation}
\boldsymbol{x}(t) = \boldsymbol{a}_5 t^5 + \boldsymbol{a}_4 t^4 + \boldsymbol{a}_3 t^3 + \boldsymbol{a}_2 t^2 + \boldsymbol{a}_1 t + \boldsymbol{a}_0,
\label{eq:quintic_polynomial}
\end{equation}
where $\boldsymbol{a}_i$ is the coefficient vector to determine the behavior of each degree.
Once initial and goal states of the trajectory $\boldsymbol{x}_{0}$ and $\boldsymbol{x}_{T}$ are given, the coefficient vector $\boldsymbol{a}_i$ can be solved by the jerk-minimization problem \cite{min-jerk-opt-1988-ICRA}. Then, the centroidal trajectory of each module from the base one can be obtained by following forward kinematics:
\begin{equation}
(\boldsymbol{r}_i,\, \boldsymbol{\alpha}_i) = f_i(\boldsymbol{r}_{\text{base}}, \, \boldsymbol{\alpha}_{\text{base}},\, \boldsymbol{q}),
\label{eq:forward_kinematics}
\end{equation}
where $f_i$ is the forward kinematics function from the base module to the $i$-th module.
The velocity and acceleration can be further calculated by differentiating \equref{eq:forward_kinematics}. Once the desired centroidal trajectory is sent to each module, the tracking control is performed independently in a distributed manner.

\subsection*{Experimental set-up}
Indoor experiments are conducted in a 8 m $\times$ 8 m $\times$ 4 m room equipped with a motion-tracking system (8 Optitrack cameras). 
The position and orientation of each robot are tracked by the motion-tracking system at a frequency of 100 Hz through 4 reflective markers attached to the airframe. 
The velocity is calculated by fusing the position from the motion-tracking system and the acceleration from the onboard IMU (200 Hz) based on Extended Kalman Filter \cite{hydrus-2021-JFR}.
This motion information is used for the flight control and contact wrench estimation.
For fully autonomous flight especially in outdoor environments, we deploy an onboard 3D LiDAR (DJI Livox MID360) and perform LiDAR Inertial Odometry \cite{fast-lio-2022-TRO} to obtain the position, orientation, and velocity instead of the motion-tracking system. 

\subsection*{Statistical Analysis}
Statistical analysis employed RMSE, box plots, and violin plots. 
RMSE is defined as $\mathrm{RMSE} = \sqrt{\frac{1}{n}\sum_{i=1}^{n} (y_i - \hat{y}_i)^2 }$. 
Violin plots are used to visualize variability across multiple trials or over multiple time steps. 
These plots are computed using the kernel density function $\hat{f}(x) = \frac{1}{n h} \sum_{i=1}^{n} K\!\left( \frac{x - x_i}{h} \right)$, where $K$ denotes the kernel function; a Gaussian kernel is adopted in this study.



\clearpage 

%
\bibliography{main} 

@article{am-survey-2022-TRO,
  author={Ollero, Anibal and Tognon, Marco and Suarez, Alejandro and Lee, Dongjun and Franchi, Antonio},
  journal={IEEE Transactions on Robotics}, 
  title={Past, Present, and Future of Aerial Robotic Manipulators}, 
  year={2022},
  volume={38},
  number={1},
  pages={626-645},
}

@article{aerila-robot-2012-IJRR,
author = {Vijay Kumar and Nathan Michael},
title ={Opportunities and challenges with autonomous micro aerial vehicles},
journal = {The International Journal of Robotics Research},
volume = {31},
number = {11},
pages = {1279-1291},
year = {2012},
}

@article{swarm-2022-SR,
  author = {Xin Zhou  and Xiangyong Wen  and Zhepei Wang  and Yuman Gao  and Haojia Li  and Qianhao Wang  and Tiankai Yang  and Haojian Lu  and Yanjun Cao  and Chao Xu  and Fei Gao },
  title = {Swarm of micro flying robots in the wild},
  journal = {Science Robotics},
  volume = {7},
  number = {66},
  pages = {eabm5954},
  year = {2022},
}

@INPROCEEDINGS{lasdra-2024-ICRA,
  author={Choe, Jaeu and Lee, Jeongseob and Yang, Hyunsoo and Nguyen, Hai-Nguyen and Lee, Dongjun},
  booktitle={2024 IEEE International Conference on Robotics and Automation (ICRA)}, 
  title={Sequential Trajectory Optimization for Externally-Actuated Modular Manipulators with Joint Locking}, 
  year={2024},
  volume={},
  number={},
  pages={8700-8706},
}

@article{cinematography-2020-JFR,
author = {Bonatti, Rogerio and Wang, Wenshan and Ho, Cherie and Ahuja, Aayush and Gschwindt, Mirko and Camci, Efe and Kayacan, Erdal and Choudhury, Sanjiban and Scherer, Sebastian},
title = {Autonomous aerial cinematography in unstructured environments with learned artistic decision-making},
journal = {Journal of Field Robotics},
volume = {37},
number = {4},
pages = {606-641},
year = {2020}
}

@article{subterran-2022-SR,
author = {Marco Tranzatto  and Takahiro Miki  and Mihir Dharmadhikari  and Lukas Bernreiter  and Mihir Kulkarni  and Frank Mascarich  and Olov Andersson  and Shehryar Khattak  and Marco Hutter  and Roland Siegwart  and Kostas Alexis},
title = {CERBERUS in the DARPA Subterranean Challenge},
journal = {Science Robotics},
volume = {7},
number = {66},
pages = {eabp9742},
year = {2022},
}

@article{drone-race-2024-TRO,
  author={Hanover, Drew and Loquercio, Antonio and Bauersfeld, Leonard and Romero, Angel and Penicka, Robert and Song, Yunlong and Cioffi, Giovanni and Kaufmann, Elia and Scaramuzza, Davide},
  journal={IEEE Transactions on Robotics}, 
  title={Autonomous Drone Racing: A Survey}, 
  year={2024},
  volume={40},
  number={},
  pages={3044-3067},
}

@INPROCEEDINGS{drone-show-2025-ICRA,
  author={Au, Tsz-Chiu},
  booktitle={2025 IEEE International Conference on Robotics and Automation (ICRA)},
  title={Contingency Formation Planning for Interactive Drone Light Shows},
  year={2025},
  volume={},
  number={},
  pages={8922-8928},
}

@article{min-snap-2011-ICRA,
  author={Mellinger, Daniel and Kumar, Vijay},
  booktitle={2011 IEEE International Conference on Robotics and Automation (ICRA)}, 
  title={Minimum snap trajectory generation and control for quadrotors}, 
  year={2011},
  volume={},
  number={},
  pages={2520-2525},
  }

@article{freestyle-2025-SR,
author = {Mingyang Wang  and Qianhao Wang  and Ze Wang  and Yuman Gao  and Jingping Wang  and Can Cui  and Yuan Li  and Ziming Ding  and Kaiwei Wang  and Chao Xu  and Fei Gao },
title = {Unlocking aerobatic potential of quadcopters: Autonomous freestyle flight generation and execution},
journal = {Science Robotics},
volume = {10},
number = {101},
pages = {eadp9905},
year = {2025},
}

@article{safe-nav-2025-SR,
author = {Yunfan Ren  and Fangcheng Zhu  and Guozheng Lu  and Yixi Cai  and Longji Yin  and Fanze Kong  and Jiarong Lin  and Nan Chen  and Fu Zhang },
title = {Safety-assured high-speed navigation for MAVs},
journal = {Science Robotics},
volume = {10},
number = {98},
pages = {eado6187},
year = {2025},
}

@article{dcad-2020-RAL,
  author={Arul, Senthil Hariharan and Manocha, Dinesh},
  journal={IEEE Robotics and Automation Letters}, 
  title={DCAD: Decentralized Collision Avoidance With Dynamics Constraints for Agile Quadrotor Swarms}, 
  year={2020},
  volume={5},
  number={2},
  pages={1191-1198},
  }

@article{omni-swarm-2022-TRO,
  author={Xu, Hao and Zhang, Yichen and Zhou, Boyu and Wang, Luqi and Yao, Xinjie and Meng, Guotao and Shen, Shaojie},
  journal={IEEE Transactions on Robotics}, 
  title={Omni-Swarm: A Decentralized Omnidirectional Visual-Inertial-UWB State Estimation System for Aerial Swarms}, 
  year={2022},
  volume={38},
  number={6},
  pages={3374-3394},
}

@article{swift-2023-nature,
  author={Kaufmann, Elia and Bauersfeld, Leonard and Loquercio, Antonio and Müller, Matthias and Koltun, Vladlen and Scaramuzza, Davide},
  journal={Nature}, 
  title={Champion-level drone racing using deep reinforcement learning}, 
  year={2023},
  volume={620},
  pages={982-987},

  }

@INPROCEEDINGS{grasp-2011-IROS,
  author={Mellinger, Daniel and Lindsey, Quentin and Shomin, Michael and Kumar, Vijay},
  booktitle={2011 IEEE/RSJ International Conference on Intelligent Robots and Systems}, 
  title={Design, modeling, estimation and control for aerial grasping and manipulation}, 
  year={2011},
  volume={},
  number={},
  pages={2668-2673},
  }

@article{soft-gripper-2024-SA,
author = {Xinyu Guo  and Wei Tang  and Kecheng Qin  and Yiding Zhong  and Huxiu Xu  and Yang Qu  and Zhaoyang Li  and Qincheng Sheng  and Yidan Gao  and Huayong Yang  and Jun Zou },
title = {Powerful UAV manipulation via bioinspired self-adaptive soft self-contained gripper},
journal = {Science Advances},
volume = {10},
number = {19},
pages = {eadn6642},
year = {2024},
}

@article{fast-grasp-2024-npjr,
  author    = {Ubellacker, Samuel and Ray, Aaron and Bern, James M. and Strader, Jared and Carlone, Luca},
  title     = {High-speed aerial grasping using a soft drone with onboard perception},
  journal   = {npj Robotics},
  year      = {2024},
  volume    = {2},
  number    = {1},
  pages     = {5},
  }

@article{am-prototype-2014-RAM,
  author={Fumagalli, Matteo and Naldi, Roberto and Macchelli, Alessandro and Forte, Francesco and Keemink, Arvid Q.L. and Stramigioli, Stefano and Carloni, Raffaella and Marconi, Lorenzo},
  journal={IEEE Robotics and Automation Magazine}, 
  title={Developing an Aerial Manipulator Prototype: Physical Interaction with the Environment}, 
  year={2014},
  volume={21},
  number={3},
  pages={41-50},
}

@article{aeroarms-2018-RAM,
  author={Ollero, Anibal and Heredia, Guillermo and Franchi, Antonio and Antonelli, Gianluca and Kondak, Konstantin and Sanfeliu, Alberto and Viguria, Antidio and Martinez-de Dios, J. Ramiro and Pierri, Francesco and Cortes, Juan and Santamaria-Navarro, Angel and Trujillo Soto, Miguel Angel and Balachandran, Ribin and Andrade-Cetto, Juan and Rodriguez, Angel},
  journal={IEEE Robotics and Automation Magazine}, 
  title={The AEROARMS Project: Aerial Robots with Advanced Manipulation Capabilities for Inspection and Maintenance}, 
  year={2018},
  volume={25},
  number={4},
  pages={12-23},
  }

@article{aam-2022-Nature,
  author    = {Zhang, Ketao and Chermprayong, Pisak and Xiao, Feng and Tzoumanikas, Dimos and Dams, Barrie and Kay, Sebastian and Kocer, Basaran Bahadir and Burns, Alec and Orr, Lachlan and Alhinai, Talib and Choi, Christopher and Darekar, Durgesh Dattatray and Li, Wenbin and Hirschmann, Steven and Soana, Valentina and others},
  title     = {Aerial additive manufacturing with multiple autonomous robots},
  journal   = {Nature},
  year      = {2022},
  volume    = {609},
  number    = {7928},
  pages     = {709--717},
}

@article{millimeter-am-2024-TRO,
  author={Wang, Meng and Chen, Zeshuai and Guo, Kexin and Yu, Xiang and Zhang, Youmin and Guo, Lei and Wang, Wei},
  journal={IEEE Transactions on Robotics}, 
  title={Millimeter-Level Pick and Peg-in-Hole Task Achieved by Aerial Manipulator}, 
  year={2024},
  volume={40},
  number={},
  pages={1242-1260},
}

@article{flyingtoolbox-2025-Nature,
  author    = {Cao, Huazi and Shen, Jiahao and Zhang, Yin and Fu, Zheng and Liu, Cunjia and Sun, Sihao and Zhao, Shiyu},
  title     = {Proximal cooperative aerial manipulation with vertically stacked drones},
  journal   = {Nature},
  year      = {2025},
  volume    = {646},
  number    = {8085},
  pages     = {576--583},
}

@INPROCEEDINGS{deform-drone-2017-IROS,
  author={Zhao, Na and Luo, Yudong and Deng, Hongbin and Shen, Yantao},
  booktitle={2017 IEEE/RSJ International Conference on Intelligent Robots and Systems (IROS)}, 
  title={The deformable quad-rotor: Design, kinematics and dynamics characterization, and flight performance validation}, 
  year={2017},
  volume={},
  number={},
  pages={2391-2396},
}

@article{foldable-drone-2019-RAL,
  author={Falanga, Davide and Kleber, Kevin and Mintchev, Stefano and Floreano, Dario and Scaramuzza, Davide},
  journal={IEEE Robotics and Automation Letters}, 
  title={The Foldable Drone: A Morphing Quadrotor That Can Squeeze and Fly}, 
  year={2019},
  volume={4},
  number={2},
  pages={209-216},
}

@article{dragon-2018-RAL,
  author={Zhao, Moju and Anzai, Tomoki and Shi, Fan and Chen, Xiangyu and Okada, Kei and Inaba, Masayuki},
  journal={IEEE Robotics and Automation Letters}, 
  title={Design, Modeling, and Control of an Aerial Robot DRAGON: A Dual-Rotor-Embedded Multilink Robot With the Ability of Multi-Degree-of-Freedom Aerial Transformation}, 
  year={2018},
  volume={3},
  number={2},
  pages={1176-1183},
}

@INPROCEEDINGS{lasdra-2018-ICRA,
  author={Yang, Hyunsoo and Park, Sangyul and Lee, Jeongseob and Ahn, Joonmo and Son, Dongwon and Lee, Dongjun},
  booktitle={2018 IEEE International Conference on Robotics and Automation (ICRA)}, 
  title={LASDRA: Large-Size Aerial Skeleton System with Distributed Rotor Actuation}, 
  year={2018},
  volume={},
  number={},
  pages={7017-7023},
}

@INPROCEEDINGS{dragonfly-2025-ICRA,
  author={Hameed, Syed Waqar and Jie, Alex Liew Jun and Imanberdiyev, Nursultan and Camci, Efe and Yau, Wei-Yun and Feroskhan, Mir},
  booktitle={2025 IEEE International Conference on Robotics and Automation (ICRA)}, 
  title={Dragonfly Drone: A Novel Tilt-Rotor Aerial Platform with Body-Morphing Capability}, 
  year={2025},
  volume={},
  number={},
  pages={8642-8648},
}

@INPROCEEDINGS{lasdra-outdoor-2019-ICRA,
  author={Park, Sangyul and Lee, Yonghan and Heo, Jinuk and Lee, Dongjun},
  booktitle={2019 International Conference on Robotics and Automation (ICRA)}, 
  title={Pose and Posture Estimation of Aerial Skeleton Systems for Outdoor Flying}, 
  year={2019},
  volume={},
  number={},
  pages={704-710},
}

@INPROCEEDINGS{chain-quadorotor-2020-ICRA,
  author={Nguyen, Huan and Dang, Tung and Alexis, Kostas},
  booktitle={2020 IEEE International Conference on Robotics and Automation (ICRA)}, 
  title={The Reconfigurable Aerial Robotic Chain: Modeling and Control}, 
  year={2020},
  volume={},
  number={},
  pages={5328-5334},
}

@article{dragon-squeeze-2020-RAL,
  author={Zhao, Moju and Shi, Fan and Anzai, Tomoki and Okada, Kei and Inaba, Masayuki},
  journal={IEEE Robotics and Automation Letters}, 
  title={Online Motion Planning for Deforming Maneuvering and Manipulation by Multilinked Aerial Robot Based on Differential Kinematics},
  year={2020},
  volume={5},
  number={2},
  pages={1602-1609},
}

@article{dargon-valve-2022-RAL,
  author={Zhao, Moju and Nagato, Keisuke and Okada, Kei and Inaba, Masayuki and Nakao, Masayuki},
  journal={IEEE Robotics and Automation Letters}, 
  title={Forceful Valve Manipulation With Arbitrary Direction by Articulated Aerial Robot Equipped With Thrust Vectoring Apparatus},
  year={2022},
  volume={7},
  number={2},
  pages={4893-4900},
  }

@article{spidar-2023-RAL,
  author={Zhao, Moju and Anzai, Tomoki and Nishio, Takuzumi},
  journal={IEEE Robotics and Automation Letters}, 
  title={Design, Modeling, and Control of a Quadruped Robot SPIDAR: Spherically Vectorable and Distributed Rotors Assisted Air-Ground Quadruped Robot}, 
  year={2023},
  volume={8},
  number={7},
  pages={3923-3930},
  }

@article{hydrus-2017-IJRR,
  author = {Moju Zhao and Koji Kawasaki and Tomoki Anzai and Xiangyu Chen and Shintaro Noda and Fan Shi and Kei Okada and Masayuki Inaba},
  title ={Transformable multirotor with two-dimensional multilinks: Modeling, control, and whole-body aerial manipulation},
  journal = {The International Journal of Robotics Research},
  volume = {37},
  number = {9},
  pages = {1085-1112},
  year = {2018},
}

@article{hydrus-2021-JFR,
author = {Zhao, Moju and Anzai, Tomoki and Shi, Fan and Maki, Toshiya and Nishio, Takuzumi and Ito, Keita and Kuromiya, Naoki and Okada, Kei and Inaba, Masayuki},
title = {Versatile multilinked aerial robot with tilted propellers: Design, modeling, control, and state estimation for autonomous flight and manipulation},
journal = {Journal of Field Robotics},
volume = {38},
number = {7},
pages = {933-966},
year = {2021}
}

@article{dragon-2023-IJRR,
author = {Moju Zhao and Kei Okada and Masayuki Inaba},
title ={Versatile articulated aerial robot DRAGON: Aerial manipulation and grasping by vectorable thrust control},

journal = {The International Journal of Robotics Research},
volume = {42},
number = {4-5},
pages = {214-248},
year = {2023},
}

@article{perching-arm-2024-TRO,
  author={Nishio, Takuzumi and Zhao, Moju and Okada, Kei and Inaba, Masayuki},
  journal={IEEE Transactions on Robotics}, 
  title={Design, Control, and Motion Planning for a Root-Perching Rotor-Distributed Manipulator}, 
  year={2024},
  volume={40},
  number={},
  pages={660-676},
}

@article{dfa-2012-IJRR,
  author={Oung, Raymond and Picallo Cruz, Miguel and D'Andrea, Raffaello},
  booktitle={2012 IEEE/RSJ International Conference on Intelligent Robots and Systems}, 
  title={A parameterized control methodology for a modular flying vehicle}, 
  year={2012},
  volume={},
  number={},
  pages={532-538},
}

@INPROCEEDINGS{modquad-2018-ICRA,
  author={Salda\~{n}a, David and Gabrich, Bruno and Li, Guanrui and Yim, Mark and Kumar, Vijay},
  booktitle={2018 IEEE International Conference on Robotics and Automation (ICRA)}, 
  title={ModQuad: The Flying Modular Structure that Self-Assembles in Midair}, 
  year={2018},
  volume={},
  number={},
  pages={691-698},
}

@INPROCEEDINGS{gripper-modquad-2018-ICRA,
  author={Gabrich, Bruno and Salda\~{n}a, David and Kumar, Vijay and Yim, Mark},
  booktitle={2018 IEEE International Conference on Robotics and Automation (ICRA)}, 
  title={A Flying Gripper Based on Cuboid Modular Robots}, 
  year={2018},
  volume={},
  number={},
  pages={7024-7030},
}

@inproceedings{h-mod-quad-2021-ICRA,
   author = {Xu, Jiawei  and Diego D{'}Antonio, S.  and Salda\~{n}a, David},
   booktitle = {2021 IEEE International Conference on Robotics and Automation (ICRA)},
   pages = {190-196},
   title = {H-ModQuad: Modular Multi-Rotors with 4, 5, and 6 Controllable DOF},
   year = {2021},
}

@article{moddessemble-2019-RAL,
   author = {Salda\~{n}a, David and Gupta, Parakh M. and Kumar, Vijay},
   issn = {23773766},
   issue = {4},
   journal = {IEEE Robotics and Automation Letters},
   month = {10},
   pages = {3402-3409},
   publisher = {Institute of Electrical and Electronics Engineers Inc.},
   title = {Design and Control of Aerial Modules for Inflight Self-Disassembly},
   volume = {4},
   year = {2019},
}

@article{modular-quadrotor-2024-RAL,
  author = {Zhang, Jihuai and Li, Fusheng and Lu, Xin and Zhang, Chaolai and Xin, Yuling and Zhao, Ruqing and Lyu, Shubin},
  journal = {IEEE Robotics and Automation Letters}, 
  title = {Design and Control of Rapid In-Air Reconfiguration for Modular Quadrotors With Full Controllable Degrees of Freedom}, 
  year = {2024},
  volume = {9},
  number = {8},
  pages = {6920-6927},
}

@article{trady-2023-AIS,
author = {Sugihara, Junichiro and Nishio, Takuzumi and Nagato, Keisuke and Nakao, Masayuki and Zhao, Moju},
title = {Design, Control, and Motion Strategy of TRADY: Tilted-Rotor-Equipped Aerial Robot With Autonomous In-Flight Assembly and Disassembly Ability},
journal = {Advanced Intelligent Systems},
volume = {5},
number = {10},
pages = {2300191},
year = {2023},
}

@article{beatle-2024-TMech,
  author={Sugihara, Junichiro and Zhao, Moju and Nishio, Takuzumi and Okada, Kei and Inaba, Masayuki},
  journal={IEEE/ASME Transactions on Mechatronics}, 
  title={BEATLE---Self-Reconfigurable Aerial Robot: Design, Control and Experimental Validation}, 
  year={2024},
  volume={},
  number={},
  pages={1-11},
}

@article{dockable-multirotor-2025-Access,
  author={Song, Yeongin and Kim, Hyunmin and Byun, Jeonghyun and Park, Keun and Kim, Murim and Lee, Seung Jae},
  journal={IEEE Access},
  title={Aerial Dockable Multirotor UAVs: Design, Control, and Flight Time Extension Through In-Flight Battery Replacement}, 
  year={2025},
  volume={13},
  number={},
  pages={96782-96799},
  }

@INPROCEEDINGS{mars-2025-ICRA,
  author={Huang, Rui and Tang, Siyu and Cai, Zhiqian and Zhao, Lin},
  booktitle={2025 IEEE International Conference on Robotics and Automation (ICRA)}, 
  title={Robust Self-Reconfiguration for Fault-Tolerant Control of Modular Aerial Robot Systems}, 
  year={2025},
  volume={},
  number={},
  pages={12614-12620},
}

@article{transport-2018-NatureP,
  author    = {Feinerman, Ofer and Pinkoviezky, Itai and Gelblum, Aviram and Fonio, Ehud and Gov, Nir S.},
  title     = {The physics of cooperative transport in groups of ants},
  journal   = {Nature Physics},
  year      = {2018},
  volume    = {14},
  number    = {7},
  pages     = {683--693},
  issn      = {1745-2481}
}

@article{army-ants-2015-NAS,
author = {Chris R. Reid  and Matthew J. Lutz  and Scott Powell  and Albert B. Kao  and Iain D. Couzin  and Simon Garnier },
title = {Army ants dynamically adjust living bridges in response to a cost–benefit trade-off},
journal = {Proceedings of the National Academy of Sciences},
volume = {112},
number = {49},
pages = {15113-15118},
year = {2015},
}

@article{modular-robot-survey-2025,
author = {Guanqi Liang and Di Wu and Yuxiao Tu and Tin Lun Lam},
title ={Decoding modular reconfigurable robots: A survey on mechanisms and design},
journal = {The International Journal of Robotics Research},
volume = {44},
number = {5},
pages = {740-767},
year = {2025},
}

@article{snail-robot-2024-NC,
  author    = {Zhao, Da and Luo, Haobo and Tu, Yuxiao and Meng, Chongxi and Lam, Tin Lun},
  title     = {Snail-inspired robotic swarms: a hybrid connector drives collective adaptation in unstructured outdoor environments},
  journal   = {Nature Communications},
  year      = {2024},
  volume    = {15},
  number    = {1},
  pages     = {3647},
}

@INPROCEEDINGS{se3-control-2010-CDC,
  author={T. {Lee} and M. {Leok} and N. H. {McClamroch}},
  booktitle={49th IEEE Conference on Decision and Control (CDC)},
  title={Geometric tracking control of a quadrotor {UAV} on {SE}(3)},
  year={2010},
  volume={},
  number={},
  pages={5420-5425}
}

@article{beetle-omni-2024-RAL,
  author={Li, Jinjie and Sugihara, Junichiro and Zhao, Moju},
  journal={IEEE Robotics and Automation Letters}, 
  title={Servo Integrated Nonlinear Model Predictive Control for Overactuated Tiltable-Quadrotors}, 
  year={2024},
  volume={9},
  number={10},
  pages={8770-8777},
}

@article{pseudo-inverse,
  author  = {Penrose, R.},
  title   = {A generalized inverse for matrices},
  journal = {Proceedings of the Cambridge Philosophical Society},
  year    = {1955},
  volume  = {51},
  pages   = {406--413}
}

@INPROCEEDINGS{momentum-based-observer-2025-ICRA,
  author={A. {De Luca} and R. {Mattone}},
  booktitle={2005 IEEE International Conference on Robotics and Automation (ICRA)},
  title={Sensorless Robot Collision Detection and Hybrid Force/Motion Control},
  year={2005},
  volume={},
  number={},
  pages={999-1004},
  }

@INPROCEEDINGS{min-jerk-opt-1988-ICRA,
  author={Kyriakopoulos, K.J. and Saridis, G.N.},
  booktitle={1988 IEEE International Conference on Robotics and Automation (ICRA)}, 
  title={Minimum jerk path generation}, 
  year={1988},
  volume={},
  number={},
  pages={364-369},
}

@book{coupled-vibration,
  author    = {Meirovitch, Leonard},
  title     = {Elements of Vibration Analysis},
  publisher = {McGraw-Hill},
  address = {New York},
  year      = {1975}
}

@INPROCEEDINGS{aerial-charging-2025-ICRA,
  author={Zhang, Ruiqi and Zhang, Dingqi and Mueller, Mark W.},
  booktitle={2025 IEEE International Conference on Robotics and Automation (ICRA)}, 
  title={ProxFly: Robust Control for Close Proximity Quadcopter Flight Via Residual Reinforcement Learning}, 
  year={2025},
  volume={},
  number={},
  pages={13683-13689},
  }

@article{fast-landing-2025-JFR,
author = {Tunney, Isaac and Bass, John and Lussier Desbiens, Alexis},
title = {Friction Shock Absorbers and Reverse Thrust for Fast Multirotor Landing on High-Speed Vehicles},
journal = {Journal of Field Robotics},
volume = {},
number = {},
pages = {},
}

@article{zmp,
title = {On the stability of anthropomorphic systems},
journal = {Mathematical Biosciences},
volume = {15},
number = {1},
pages = {1-37},
year = {1972},
issn = {0025-5564},
author = {M. Vukobratovi\'{c} and J. Stepanenko},
}

@article{fast-lio-2022-TRO,
  author={Xu, Wei and Cai, Yixi and He, Dongjiao and Lin, Jiarong and Zhang, Fu},
  journal={IEEE Transactions on Robotics}, 
  title={FAST-LIO2: Fast Direct LiDAR-Inertial Odometry}, 
  year={2022},
  volume={38},
  number={4},
  pages={2053-2073},
}

@patent{switchable-permanent-magnet-patent,
  author       = {Kocijan, F. and Underwood, P. J.},
  title        = {Switchable (variable) permanent magnet device},
  number       = {AUPQ446699A0},
  year         = {2000},
  date         = {2000-01-06},
  country      = {AU},
  assignee     = {Kocijan, F.; Underwood, P. J.},
  filing_date  = {1999-12-06},
  url          = {https://patents.google.com/patent/AUPQ446699A0}
}

@misc{fire-ant,
  title={Solenopsis xyloni 248007479.jpg (2022)},
  note = {\url{https://commons.wikimedia.org/wiki/File:Solenopsis_xyloni_248007479.jpg}},
}

@misc{ant-carrying,
  title={Red ant carrying food to nest},
  note = {https://stock.adobe.com/jp/video/red-ant-carrying-food-to-nest/59250838},
}

@misc{ant-raft,
  title={Fire ants create raft out of other ants to live through flood},
  note = {https://stock.adobe.com/jp/video/fire-ants-create-raft-out-of-other-ants-to-live-through-flood/384012267},
}

@misc{ant-bridge,
  title={Ant action standing.Ant bridge unity team,Concept team work together Red ant,Weaver Ants (Oecophylla smaragdina),Action of ant carry food},
  note = {https://stock.adobe.com/jp/video/ant-action-standing-ant-bridge-unity-team-concept-team-work-together-red-ant-weaver-ants-oecophylla-smaragdina-action-of-ant-carry-food/420082761},
}

%
%
%
%
%
%



\newpage


\renewcommand{\thefigure}{S\arabic{figure}}
\renewcommand{\thetable}{S\arabic{table}}
\setcounter{figure}{0}
\setcounter{table}{0}
\setcounter{equation}{0}
\setcounter{page}{1} 


\begin{center}
\section*{Supplementary Information for Self-assembling Modular Aerial Robot for Versatile Aerial Tasks}

J.~Sugihara,
M.~Kitagawa,
J.~Li,
Y.~Li,
T.~Nishio,
K.~Okada,
M.~Zhao$^{\ast}$\\
\small$^\ast$Corresponding author. Email: chou@dragon.t.u-tokyo.ac.jp\\
\end{center}

\subsubsection*{This PDF file includes:}
Supplementary Results\\
Supplementary Methods\\
Figures S1 to S4\\
Tables S1 to S3\\

\newpage


\subsection*{Supplementary Results}
\subsubsection*{Performance comparison of morphing with and without contact wrench control}
\begin{figure}
 \centering
 \includegraphics[width=1.0\textwidth]{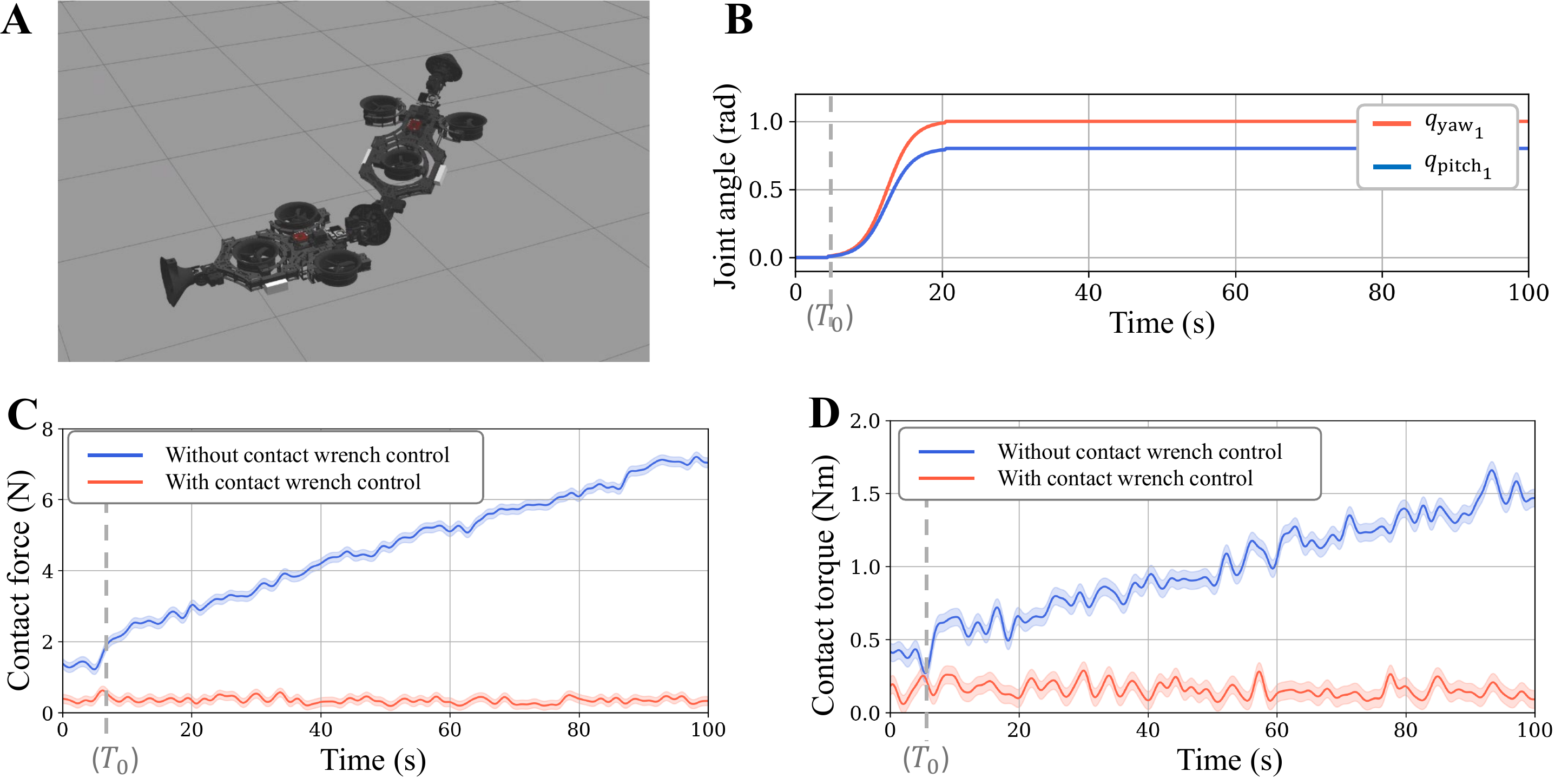}
 \caption{\textbf{Simulation evaluation of contact wrench control.}
(\textbf{A}) Comparison of the stability of the morphing motion by the two assembled modules with and without the proposed contact wrench control. The flight simulations were conducted in Gazebo.
(\textbf{B}-\textbf{D}) Time histories of the target joint angles during morphing (B), the contact forces between modules (C), and the contact torques between modules (D). The solid lines in (C) and (D) represent values processed by a low-pass filter with a cutoff frequency of 5 Hz, and the shaded regions denote the standard deviation over the entire interval. In (B-D), $T_{0}$ denotes the time at which the joint motion began.
}
 \label{fig:ex_fig2}
\end{figure}

Simulations of two-module morphing motion were conducted both with and without the contact wrench control (\figref{fig:ex_fig2}). To minimize the gap between the simulation and the reality, Gaussian noise with a standard deviation of $1{\times}10^{-3}$ was added to the centroidal pose. 
The desired contact wrench was zero ($\bar{\mathbf{w}}^{\mathrm{ctc}}_i = \boldsymbol{0}$ in \equref{eq:ff_dynamics}), so the actual contact wrench should be also around zero.
At $T_o$, the target joint angles started to change from 0 to 0.8 rad. When the contact wrench control was applied, the contact wrench immediately converged to zero even during joint motion.
In contrast, when the contact wrench control was disabled, the inter-module contact wrench continuously increased due to the reaction torque of the joint actuators.
This excessive contact wrench also happened in reality, and caused joint overload, thrust saturation, and instability during disassembly which was discussed in our previous work \cite{beatle-2024-TMech}.

\subsubsection*{Scalability of multi-connected control framework}
\begin{figure}
 \centering
 \includegraphics[width=1.0\textwidth]{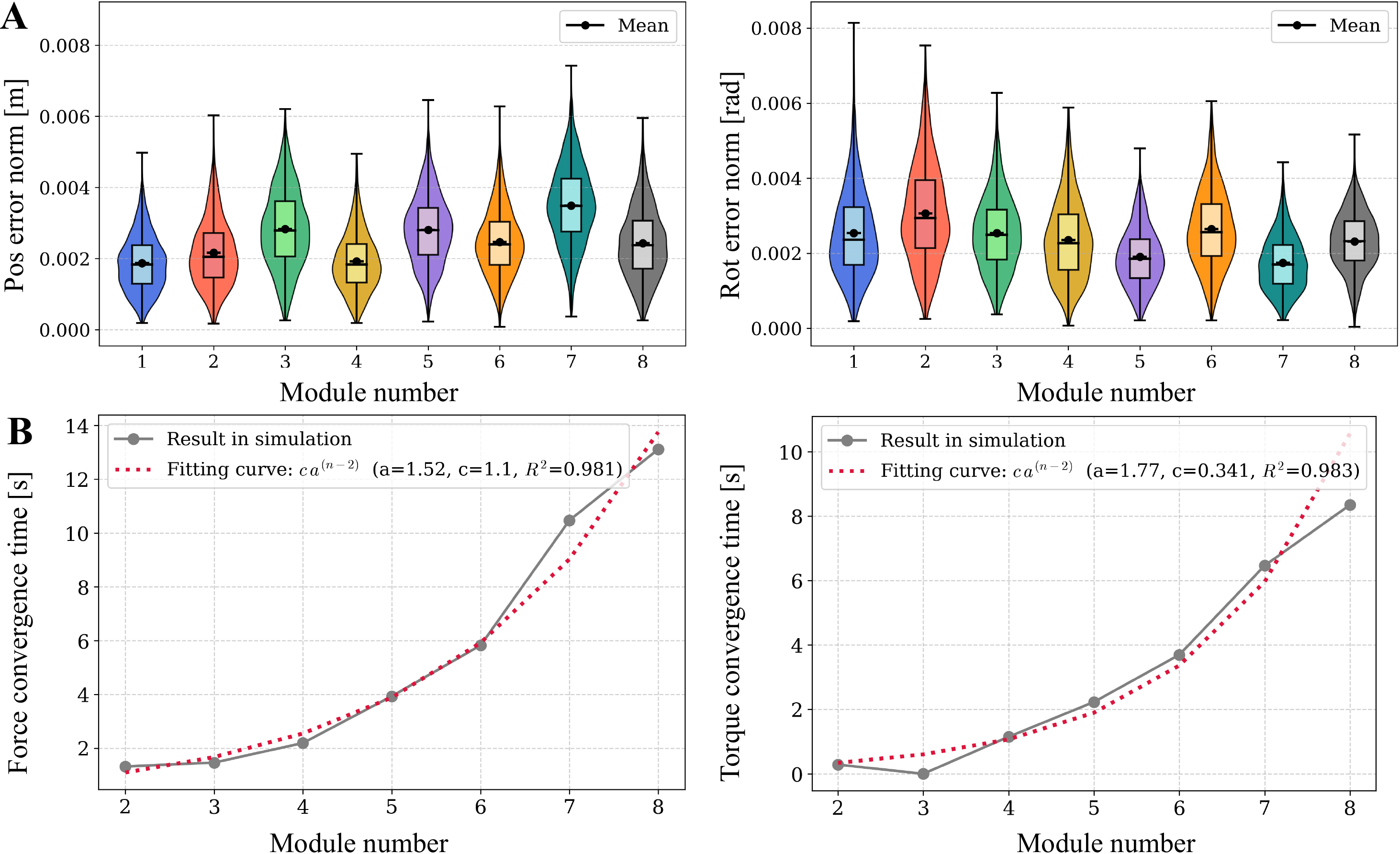}
 \caption{\textbf{Simulation evaluation of modular scalability.}
(\textbf{A}) Comparison of the positional and rotational errors in the assembled mode with different numbers of modules (from 1 to 8).
(\textbf{B}) Comparison of the convergence time of the contact wrench control during the transition from takeoff to steady flight.
}
 \label{fig:ex_fig3}
\end{figure}

We also evaluated the relationship between the number of assembled modules and flight stability (\figref{fig:ex_fig3}).
The simulation results showed that the position and attitude errors remained almost constant regardless of the number of assembled modules, which corresponded to the behavior in the real experiment (\figref{fig:docking_morphing}D).
On the other hand, the convergence time of contact wrench from the start of docking increased exponentially with the number of assembled modules. This exponential trend is consistent with the convergence analysis discussed in the ``Supplementary Methods'' Convergence of the control flow for multi-connected flight.

\subsubsection*{Hardware efficiency analysis}
Although optimization of energy efficiency was not a primary objective of this study, several key hardware parameters of LEGION were compared with those of previous self-reconfigurable aerial robot systems. 
\tabref{tab:a-msrr} summarizes a comparison between the state-of-the-art sele-assembly or self-disassembly platforms and LEGION. 
LEGION is the only system capable of fully autonomous self-assembly, self-disassembly, and morphing flight, while also incorporating articulated joints that are absent in other systems. 
Despite this additional mechanical complexity, LEGION achieved superior performance in both flight endurance and the mass ratio of the docking mechanism, $D_{\text{mass}}$ (i.e., the proportion of the coupling mechanism’s mass to the total module mass), compared with major existing platforms. 

\subsection*{Supplementary Methods}
\subsubsection*{Task metrics}
The metrics for ``pushing'' and ``pulling'' (\figref{fig:manipulation_results}) are defined directly by the total exertable force from the whole configuration:
\begin{equation}
V_{\text{push}} = V_{\text{pull}} = n.
\end{equation}

For ``rotating'',  swarm configuration follows $O(n)$ scalability with respect to the number of units, similar to ``pushing'', and ``pulling''. In contrast, both LEGION and multi-link configuration can transmit internal forces across links, allowing the exertable end-effector torque to scale as $O(n^2)$. Therefore, the metrics are:
\begin{equation}
V^{\text{swarm}}_{\text{rot}}=n,\;\;V^{\text{multi}}_{\text{rot}} = V^{\text{legion}}_{\text{rot}}=n^2.
\end{equation}

For ``grasping'', the metric is defined based on the reachable workspace of the end-effector in multi-link configuration. Assuming that each inter-link joint provides two rotational degrees of freedom, the reachable region expands approximately spherically with respect to the base link, which can be defined as a cubic growth $O(n^3)$ :
\begin{equation}
V^{\text{legion}}_{\text{grasp}}=V^{\text{multi}}_{\text{grasp}} = n^3,\;\; V^{\text{swarm}}_{\text{grasp}} = 1.
\end{equation}
Note that for swarm configuration, grasping cannot be defined in the first place. A nominal value is assigned here solely for comparative purposes.

The performance for ``carrying'' is evaluated based on the stability of object support. We adopt the convex hull area formed by the support points---commonly used for stability analysis in legged robots \cite{zmp}---as the stability metric. Assuming that the $n$ support points form a regular $n$-gon with a fixed edge length, the convex hull area scales as $O(n^2)$ with the number of support points. For multi-link configuration, however, the number of end-effectors (i.e., support points) does not increase with the number of links; thus, the convex hull area remains $O(1)$. Accordingly, the metric for ``carrying'' is:
\begin{equation}
V^{\text{legion}}_{\text{carry}} = V^{\text{swarm}}_{\text{carry}} = n^2,\;\;
V^{\text{multi}}_{\text{carry}}=
\begin{cases}
1 & (n = 1), \\
4 & (n \ge 2).
\end{cases}
\end{equation}

Smaller configuration is capable of maneuvering in more cluttered environments, which corresponds to nimbleness.
In LEGION and the swarm configuration, the physical size of each module or agent does not change as system size increases. In contrast, multi-link configuration grows in overall size proportionally to $O(n)$ as the number of links increases. Therefore, the metric for ``nimbleness'' for each configuration can be defined as:
\begin{equation}
V^{\text{multi}}_{\text{nimble}}=\frac{1}{n},\;\;V^{\text{legion}}_{\text{nimble}} = V^{\text{swarm}}_{\text{nimble}} = 1.
\end{equation}

\tabref{tab:morphing} presents a comparison between LEGION and other multi-link aerial robots capable of full-body morphing. 
LEGION, DRAGON~\cite{dragon-2018-RAL}, SPIDAR~\cite{spidar-2023-RAL}, and LASDRA~\cite{lasdra-2024-ICRA}, all of which enable three-dimensional morphing, employ two-degree-of-freedom vectored ducted-fan units. 
In contrast, HYDRUS~\cite{hydrus-2017-IJRR} and ModQuad Grasp~\cite{gripper-modquad-2018-ICRA} which are limited to planar (2D) morphing, uses large open-propeller configurations. 
The rotor arrangements adopted in 3D morphing systems~\cite{dragon-2018-RAL,spidar-2023-RAL,lasdra-2024-ICRA} provide higher maximum thrust and comparable payload capacities, whereas those in 2D morphing systems~\cite{hydrus-2017-IJRR, gripper-modquad-2018-ICRA} offer lower maximum thrust but higher propulsive efficiency and longer continuous flight times. 
As discussed in the ``Discussion'' section, these differences represent inherent design trade-offs governed by the current technological limits of fundamental components such as motors and batteries.

\subsubsection*{Convergence of the control flow for multi-connected flight}
For the contact wrench control in multi-connected flight to converge, convergence of the wrench observer must first be ensured.  
By differentiating \equref{eq:wrench_observer} with respect to time, the following expression is obtained:
\begin{equation}
\dot{\hat{\mathbf{w}}}^{\text{ext}} 
+ K_I\hat{\mathbf{w}}^{\text{ext}} 
= K_I\mathbf{w}^{\text{ext}}.
\label{eq:observer_lpf}
\end{equation}
Let the estimation error be defined as $\boldsymbol{e}_{obs} = \hat{\mathbf{w}}^{\text{ext}} - \mathbf{w}^{\text{ext}}$.
Then \equref{eq:observer_lpf} can be rewritten as:
\begin{equation}
\dot{\boldsymbol{e}}_{obs} = -K_I \boldsymbol{e}_{obs} - \dot{\mathbf{w}}^{\text{ext}}.
\end{equation}
Since $\dot{\mathbf{w}}^{\text{ext}}$ is bounded, the error system is input-to-state stable (ISS), and $\|\boldsymbol{e}_{obs}(t)\|$ remains bounded with an ultimate bound proportional to $\sup_{t\ge 0}\|\dot{\mathbf{w}}^{\text{ext}}(t)\|$.
In particular, if $\dot{\mathbf{w}}^{\text{ext}}(t)\to 0$ (e.g., $\mathbf{w}^{\text{ext}}$ becomes constant),
then $\boldsymbol{e}_{obs}(t)\to 0$ and the observer asymptotically converges.
Because the convergence of the contact wrench estimator defined by \equref{eq:sim_eq_ctc} depends on that of the external wrench observer, it also inherits this property.
As described in the ``Contact wrench control'', when the contact wrench estimation is accurate, each module’s contact wrench converges to its desired value within one control step. 
In practice, however, estimation errors, communication delays, and coordinate transformation inaccuracies exist.  
Defining the contact wrench error as 
$\boldsymbol{e}_i(t) := {}{^\mathrm{des}}\mathbf{w}^{\mathrm{ctc}}_i(t) - \mathbf{w}^{\mathrm{ctc}}_i(t)$,  
the discrete-time dynamics can be expressed as:
\begin{equation}
\boldsymbol{e}(k+1)=R(k)\,\boldsymbol{e}(k)+S\,\varepsilon(k),
\label{eq:S8_original}
\end{equation}
where $R(k)=R\!\left(q(k),\delta\right)$ is time-varying through the robot configuration $q(k)$ and discretization $\delta$, and $\varepsilon(k)$ denotes the (bounded) exogenous perturbation term. 
Under the approximation that, due to the smoothing coefficient $\gamma$ introduced in \equref{eq:ff_smoothing}, the error is filtered by the same first-order smoothing law,
\begin{equation}
\boldsymbol{e}^{*}(k+1)=\gamma\,\boldsymbol{e}(k+1)+(1-\gamma)\,\boldsymbol{e}^{*}(k),
\label{eq:error_smoothing}
\end{equation}
and substituting \equref{eq:S8_original} into \equref{eq:error_smoothing} yields the smoothed error dynamics
\begin{equation}
\boldsymbol{e}^{*}(k+1)=R_{\gamma}(k)\,\boldsymbol{e}^{*}(k)+\gamma S\,\varepsilon(k),
\qquad
R_{\gamma}(k):=(1-\gamma)I+\gamma R(k).
\label{eq:S8_smoothed}
\end{equation}
where $I$ represents an identify matrix. Assume there exists a constant $\alpha\in[0,1)$ such that
\begin{equation}
\sup_{k\ge 0}\,\bigl\|R_{\gamma}(k)\bigr\| \;\le\; \alpha \;<\; 1,
\label{eq:uniform_contraction}
\end{equation}
for a chosen induced matrix norm $\|\cdot\|$.
Then \equref{eq:S8_smoothed} is uniformly contractive and admits the bound
\begin{equation}
\bigl\|\boldsymbol{e}^{*}(k)\bigr\|
\le
\alpha^{k}\bigl\|\boldsymbol{e}^{*}(0)\bigr\|
+
\sum_{i=0}^{k-1}\alpha^{k-1-i}\,\gamma\|S\|\,\|\varepsilon(i)\|.
\label{eq:ek_bound}
\end{equation}
If $\varepsilon(k)$ is bounded, i.e.\ $\sup_{k\ge 0}\|\varepsilon(k)\|<\infty$, it follows that
\begin{equation}
\limsup_{k\to\infty}\bigl\|\boldsymbol{e}^{*}(k)\bigr\|
\le
\frac{\gamma\|S\|}{1-\alpha}\;
\sup_{k\ge 0}\|\varepsilon(k)\|.
\label{eq:S9_revised}
\end{equation}
Therefore, under \equref{eq:uniform_contraction} and bounded $\varepsilon(k)$, the smoothed contact-wrench error $\boldsymbol{e}^*(k)$ is ultimately bounded and converges to a neighborhood of the origin whose size is given by \equref{eq:S9_revised}.
Condition \equref{eq:uniform_contraction} is a sufficient (norm-based) stability condition for the time-varying system \equref{eq:S8_smoothed}. In the time-invariant special case $R(k)\equiv R$, \equref{eq:uniform_contraction} reduces to $\| (1-\gamma)I+\gamma R\|<1$, and one may equivalently state $\rho((1-\gamma)I+\gamma R)<1$ using the spectral radius $\rho(\cdot)$.

In practical implementation, the sufficient contraction condition is enforced using an induced norm.
Specifically, we aim to keep $\sup_k \|R_\gamma(k)\| < 1$, where $R_\gamma(k)=(1-\gamma)I+\gamma R(k)$.
Since $R(k)$ (and hence $R_\gamma(k)$) is composed of inter-module wrench transformation matrices, its norm may grow rapidly with the number of connected modules.
Accordingly, the smoothing coefficient $\gamma$ is decreased with $n$ to maintain the contractivity of $R_\gamma(k)$.
Accordingly, the smoothing coefficient $\gamma$ is defined as:
\begin{equation}
\gamma_n = \frac{\gamma_2}{2^{n-2}},
\end{equation}
where $\gamma_2$ is the tuned coefficient for the two-module configuration, set to $\gamma_2 = 0.4$ in the implemented system.

\clearpage
\begin{figure}
 \centering
 \includegraphics[width=1.0\textwidth]{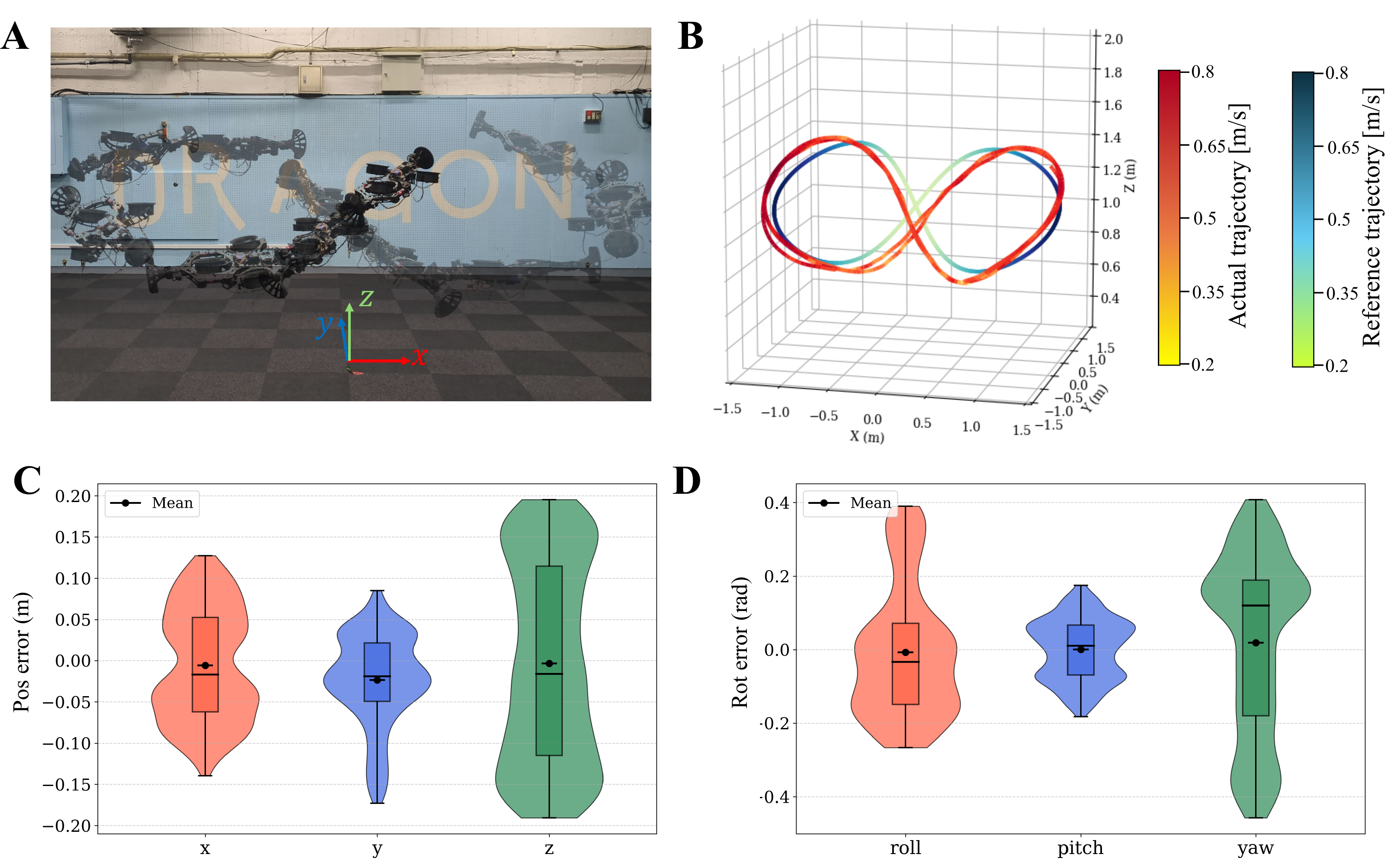}
 \caption{\textbf{Individual nimble flight.}
(\textbf{A}) LEGION is capable of tracking omni-directional trajectory that requires independent control for translational and rotational motion. 
(\textbf{B}) Target trajectory and actual trajectory with the robot velocity at each point shown by a colormap. In this experiment, LEGION completed three laps of the target trajectory. 
(\textbf{C} and \textbf{D}) Norms of the positional error (C) and rotational error (D) during trajectory tracking.
}
 \label{fig:ex_fig1}
\end{figure}

\begin{figure}
 \centering
 \includegraphics[width=1.0\textwidth]{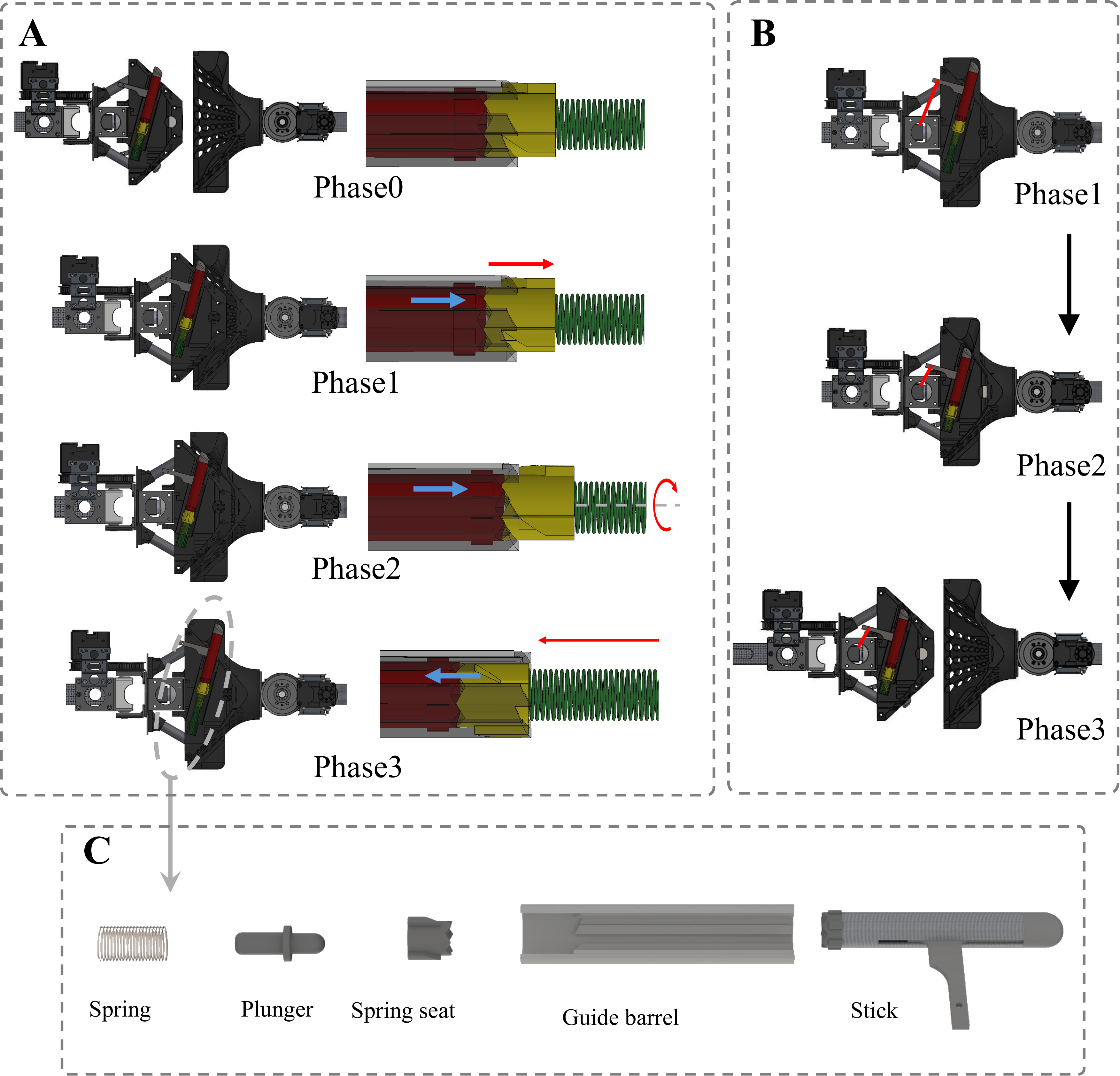}
 \caption{\textbf{Detail of hardware design.}
(\textbf{A}) Operating process of the retractable mechanism during docking. The two interfaces first come into contact, and the stick on the male side is ``clicked'' (Phase 1). Once the stick reaches the bottom, the yellow spring seat rotates (Phase 2), and the spring ejects the stick outward along the guide barrel (Phase 3). As soon as Phase 3 begins, the stick is aligned with and inserted into the receptor hole of the female side, thereby completing the interlocking.
(\textbf{B}) Operating process during separation. For separation, a wire driven by a spool and a servo motor mounted on the male side pulls the stick back into the mechanism.
(\textbf{C}) Exploded view of the retractable mechanism. 
}
 \label{fig:ex_fig4}
\end{figure}

\clearpage 

\begin{table} 
	\centering
	\caption{\textbf{Specifications of self-reconfigurable modular aerial robots}}
	\label{tab:a-msrr} 
	\begin{tabular}{lccccc} 
         \\
         Robots & Self-assembly & Self-disassembly & Morphing &\shortstack{Flight endurance\\ {[min]}} & \shortstack{$D_{\text{mass}}$ \\ $= m_{\text{dock}} / m_{\text{total}}$}\\
         \hline
         \hline
         LEGION(ours) & yes & yes & yes & 6 & 0.12\\
         BEATLE\cite{beatle-2024-TMech} & yes & yes & no & 6 & 0.15 \\
         TRADY\cite{trady-2023-AIS} & yes & yes & no & 8 & 0.15 \\
         AirDock\cite{dockable-multirotor-2025-Access} & yes & yse & no & 4 & no data\\
         ModQuad 2018\cite{modquad-2018-ICRA} & yes & no & no & 5& 0.21 \\
         ModQuad 2019\cite{moddessemble-2019-RAL} & no data & yes & no & no data & 0.18 \\
         \hline
	\end{tabular}
\end{table}

\begin{table} 
	\centering
	\caption{\textbf{Specifications of multilinked morphing aerial robots}}
	\label{tab:morphing} 
	\begin{tabular}{lccccc} 
         \\
         Robots & \shortstack{Morphing \\ Dimension} & \shortstack{Number \\ of Links} & \shortstack{Rotor \\ Configuration} &\shortstack{Payload\\ {[N]}} & \shortstack{Flight endurance\\ {[min]}}\\
         \hline
         \hline
         LEGION & 3D & n & 2DoF Vectoring & 24$\times$n & 6 \\
         DRAGON\cite{dragon-2018-RAL} & 3D & 4 & 2DoF Vectoring & 94 & 3 \\
         SPIDAR\cite{spidar-2023-RAL} & 3D & 9 & 2DoF Vectoring& 184 & 9 \\
         LASDRA\cite{lasdra-outdoor-2019-ICRA} & 3D & 4 & Fixed &no data & no data\\
         DragonFly\cite{dragonfly-2025-ICRA} & 2D & 4 & 1DoF Vectoring &no data& no data\\
         HYDRUS\cite{hydrus-2017-IJRR} & 2D & 4 & Fixed & 30 & 15 \\
         ModQuad Grasp\cite{gripper-modquad-2018-ICRA} & 2D & 4 & Fixed & $\ge$ 0.14  &$\le$ 7  \\
         \hline
	\end{tabular}
\end{table}

\begin{table} 
	\centering
	\caption{\textbf{LEGION's components specifications}}
	\label{tab:bom} 
	\begin{tabular}{lll} 
         \\
         Component & Reference & Specifications\\
         \hline
         \hline
         Thrust motor & Cobra 2217/12 & 70 g, 1550 Kv\\
         Rotor vectoring Motor & Dynamixel XL-430-W250 & 56 g, 1.5 Nm\\
         Joint Motor & Dynamixel XH-430-w210 & 82 g, 2.5 Nm\\
         Battery & \shortstack{Turnigy Heavy Duty \\ 3000mAh 6S LiPo} & 446 g, 60C discharge\\
         Propeller & GEMFAN 5055S-3 & \\
         ESC & T-MOTOR F55A & \\
         Microcontroller unit & STM32H743 & 480 MHz, ARM Cortex-M7\\
         IMU & ICM-2048 & range: $\pm 2000^\circ$/s, $\pm 16$ g\\
         LiDAR & Livox MID360& 265 g, range: 70 m with 80 \% reflectivity\\
         Onboard Computer & Khadas VIM4 & 2.2 GHz Quad core ARM Cortex-A73\\
         \hline
	\end{tabular}
\end{table}




\clearpage 




\end{document}